\documentclass[10pt,twocolumn,letterpaper]{article}

\usepackage{cvpr}
\usepackage{times}
\usepackage{epsfig}
\usepackage{graphicx}
\usepackage{amsmath}
\usepackage{amssymb}

\usepackage[pagebackref=true,breaklinks=true,letterpaper=true,colorlinks,bookmarks=false]{hyperref}
\usepackage{algorithm}
\usepackage[noend]{algpseudocode}
\usepackage{multirow,graphicx,paralist,bigdelim}
\usepackage{booktabs}
\usepackage{pifont} 
\usepackage{array}
\newcolumntype{L}[1]{>{\raggedright\let\newline\\\arraybackslash\hspace{0pt}}m{#1}}
\newcolumntype{C}[1]{>{\centering\let\newline\\\arraybackslash\hspace{0pt}}m{#1}}
\newcolumntype{R}[1]{>{\raggedleft\let\newline\\\arraybackslash\hspace{0pt}}m{#1}}

\renewcommand{\vec}[1]{\mathbf{#1}}
\usepackage[pagebackref=true,breaklinks=true,letterpaper=true,colorlinks,bookmarks=false]{hyperref}

\cvprfinalcopy 

\def\cvprPaperID{7459} 

\ifcvprfinal\fi
\begin{document}

\title{Revisiting Saliency Metrics: Farthest-Neighbor Area Under Curve}

\author{Sen Jia\\
Ryerson University\\
{\tt\small sen.jia@ryerson.ca}
\and
Neil D. B. Bruce\\
Ryerson University, Vector Institute\\
{\tt\small bruce@ryerson.ca}
}

\maketitle
\begin{abstract}
Saliency detection has been widely studied because it plays an important role in various vision applications, but it is difficult to evaluate saliency systems because each measure has its own bias. In this paper, we first revisit the problem of applying the widely used saliency metrics on modern Convolutional Neural Networks(CNNs). Our investigation shows the saliency datasets have been built based on different choices of parameters and CNNs are designed to fit a dataset-specific distribution. Secondly, we show that the Shuffled Area Under Curve(S-AUC) metric still suffers from spatial biases. We propose a new saliency metric based on the AUC property, which aims at sampling a more directional negative set for evaluation, denoted as Farthest-Neighbor AUC(FN-AUC). We also propose a strategy to measure the quality of the sampled negative set. Our experiment shows FN-AUC can measure spatial biases, central and peripheral, more effectively than S-AUC without penalizing the fixation locations. Thirdly, we propose a global smoothing function to overcome the problem of few value degrees (output quantization) in computing AUC metrics. Comparing with random noise, our smooth function can create unique values without losing the relative saliency relationship. 
\end{abstract}
\vspace{-0.5cm}

\section{Introduction}
Extensive studies have been proposed to predict the most salient region within an image. Saliency methods can be roughly grouped into two categories, bottom-up and top-down. The former considers the visual stimuli of an image to determine the Regions Of Interest(ROIs); while the latter one assumes the ROI is task-dependent, prior knowledge plays a significant part in saliency prediction. With the recent development of Convolutional Neural Networks (CNNs), saliency prediction heavily relies on model-based algorithms which can be trained in an end-to-end fashion. A new question has emerged regarding what type of saliency features are the best design for applications, top-down or bottom-up? Hand-crafted or CNN-learned?

\begin{figure}[t]
	\setlength\tabcolsep{1.0pt}
	\centering
	\resizebox{.4\textwidth}{!}{
		\begin{tabular}{*{3}{c }}
			\includegraphics[width=0.15\textwidth]{imgs/sample/img_1.png}&
			\includegraphics[width=0.15\textwidth]{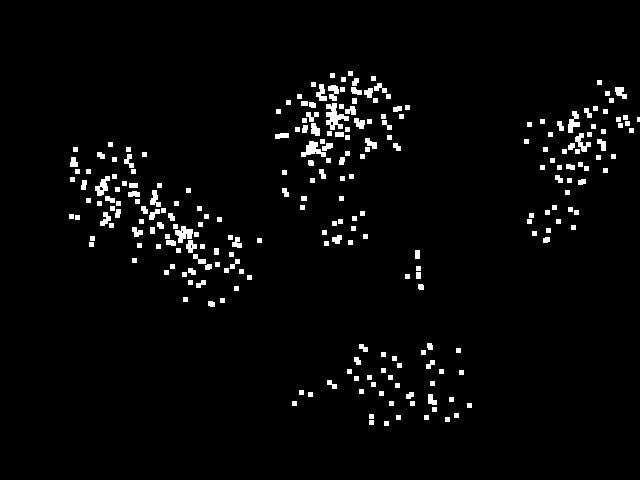}&
			\includegraphics[width=0.15\textwidth]{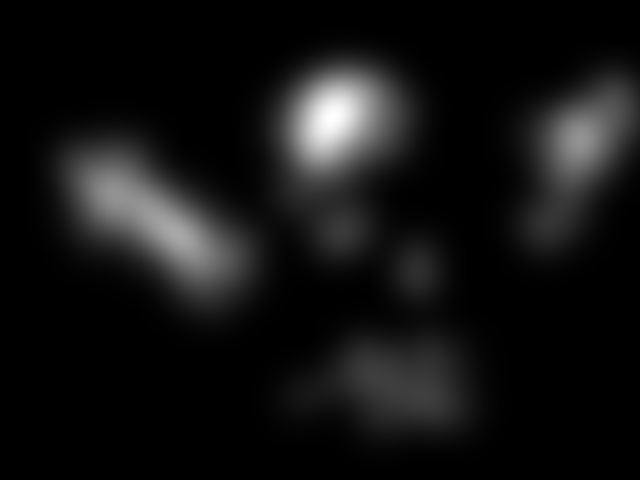}\\
			Image & Fixation & Distribution
		\end{tabular}}
		\caption{A saliency sample from the SALICON dataset.}
		\label{fig:sample}
		\vspace{-0.7cm}
	\end{figure}

The measurement of saliency is still challenging because the definition of ``saliency'' varies depending on the vision task \cite{Boisvert16}, so saliency algorithms can also be grouped by different taxonomies for different purposes \cite{Bylinskii15,Frintrop10,Borji13b}. In this paper, we follow the common problem setting in \cite{MIT1003,SPROC}, computational models are trained to predict the most salient region to the Human Visual System(HVS). Not only is this a common assumption to all the CNN-based methods \cite{SALICON,TORONTO,MIT300,DEEPGAZE2,SAMRes,MLNet,DEEPGAZE1,PDP}, but it also can be used in an extremely wide range of applications \cite{Zund13,Ancuti11,Zhang16,Wang17,Jia17b,Hu18,Li18}. However, it is still difficult to comprehensively evaluate saliency models due to the bias of each metric. For instance, the challenge of Large-Scale Scene Understanding \cite{SALICON} uses seven saliency metrics, Shuffled Area Under Curve (S-AUC), Information Gain (IG) \cite{Matthias15}, Normalized Scanpath Saliency (NSS) \cite{NSS},  Pearson’s Correlation Coefficient (CC), AUC-Judd, SIMilarity (SIM) and Kullback–Leibler Divergence (KLD) \cite{KL}. Another public saliency benchmark, MIT300 \cite{MIT300}, applies eight saliency metrics, AUC-Judd, AUC-Borji, S-AUC, NSS, CC, KLD, SIM and  Earth Mover’s Distance (EMD) \cite{EMD}. The use of multiple measures indicates that it is difficult to evaluate a model from only one angle. Previous studies intended to categorize and compare those metrics, e.g., \cite{Riche13,Bylinskii19} grouped the metrics into location-based and distribution-based. As shown in Figure~\ref{fig:sample}, the location-based measure consists of a set of fixation locations captured by an eye tracker or mouse click. While the distribution of saliency is normally considered as a post-process on the raw data by applying a Gaussian filter.  

To evaluate a saliency model, one solution is to overcome the ``disagreement'' among the metrics. Kummerer \textit{et al.} \cite{Matthias15} proposed to optimize the saliency scale, center bias and spatial blurring jointly. But their post-process can hardly satisfy all the metrics simultaneously, the process requires all the compared models and optimization only uses the loss of IG. Later this idea was extended to a metric-tailored design \cite{Matthias18}, saliency models and saliency maps are decoupled so that one model can output different metric specific maps. However, their solutions are based on the assumption that all the metrics are able to evaluate saliency reasonably, the output should be optimized separately and specifically for each metric. In this work, we are more interested in investigating the differences among those metrics in theory. We believe some of the metrics may contain inherent drawbacks so that not necessarily all the metrics should be considered. The first contribution of this work is that we revisit the widely used saliency metrics based on \cite{Bylinskii19}, showing that the ``balanced'' metrics, NSS and CC, still have limitations in evaluating modern CNN-based systems. Saliency datasets have been created using their own choices of Gaussian standard deviation and a CNN model learns to fit this biased distribution, see Section~\ref{sec:metric}. 

Center bias is a long-standing problem in saliency evaluation, simply placing a Gaussian distribution at the center could outperform a well-designed system on most of the metrics \cite{SUN}. S-AUC is a common solution used by the SALICON and the MIT benchmarks, a negative set is sampled based on the positive from other images within the same dataset. The study \cite{Bylinskii19} shows a centered Gaussian distribution can only achieve an S-AUC score of $0.5$. However, S-AUC has a strong bias that it only considers negatives near the center, peripherally-favored systems can achieve higher scores \cite{Bruce15}. In this paper, we propose a new saliency metric which introduces one more constraint on the spatial relationship between the positive and negative. We show that the distribution of all fixations can be interpreted as a 2D probability density distribution. Our method builds the negative set for each image by searching its farthest neighbors according to the distribution similarity, denoted as FN-AUC. We also propose a fast version of our FN-AUC in case the size of the dataset is too large. To compare with S-AUC, we propose a strategy to measure the quality of the sampled negative set, which takes the spatial relationship of both the center bias and the positive into account, see Section~\ref{sec:fnauc}. Our experiment shows FN-AUC can draw a more reasonable negative set in order to penalize the center bias only without undermining the true positives, Section~\ref{sec:case}.

Another contribution of this work is that we propose a global smoothing strategy in computing AUC metrics. Based on the MIT benchmark, it is problematic to compute the receiver operating characteristic (ROC) property when many locations share the same value magnitude, which could result in a lower performance as well. This is a scenario that is common for CNN based models which can produce near binary outputs. One solution could be jittering a map by adding small random noise, but this may break the relative saliency rank. We propose to apply a Gaussian filter using a relatively large standard deviation, the output is expected to cover all regions within a map. In this way, our method can generate a map with unique values for an AUC metric, meanwhile retaining the relative saliency relationship, see Section~\ref{sec:global}.

\section{Revisiting Saliency Metrics}\label{sec:metric}
In this section, we first revisit the widely used saliency metrics based on previous studies \cite{Riche13,Bruce15,Bylinskii19}. We further investigate the impact of applying the balanced saliency metrics across datasets, NSS and CC.

\begin{table}[t]
\begin{center}
\resizebox{.4\textwidth}{!}{
\begin{tabular}{|c|c|c|c|}
\hline
Dataset & Size (Width by Height) & $\sigma$ & CC \\ 
\hline
Toronto~\cite{TORONTO} & $681 \times 511$ & 20 & .998 \\
MIT1003~\cite{MIT1003} & (Min - Max)$405 - 1024$ & 24 & .998 \\
CAT2000~\cite{CAT2000} & $1920 \times 1080$ & 41 & .998 \\
SALICON~\cite{SALICON} & $640 \times 480$ & 19 & .999 \\
\hline
\end{tabular}
}
\end{center}
\caption{The attributes of the four saliency datasets. Gaussian processes with different standard deviations are used to generate the distribution ground-truth.}
\label{tab:dataset}
\vspace{-0.6cm}
\end{table}

\subsection{Distribution-based Metrics}
The distribution-based saliency metrics, SIM, CC, EMD and KLD, consider each saliency map as a distribution then the similarity between two maps can be measured based on a probabilistic view. The drawbacks of IG, SIM and KLD have already been studied in that they focus more on FNs than FPs\footnote{False Postive(FP), False Negative(FN), True Positive(TP), True Negative(TN).} which leads to a biased evaluation. While the EMD metric is sensitive to the sparsity of the map, a lower score can be obtained due to fewer bins requiring moving. The CC measure is recommended for penalizing FPs and FNs equally \cite{Bylinskii19}. However, when comparing two distributions, the ``shape'' of each distribution also plays an important role (the choice of the Gaussian sigma $\sigma$). Especially for CNN-based saliency systems, a CNN model is designed to learn the distribution information from the training set, a lower performance may be achieved only because the test set is drawn from a different distribution. This is a common problem in practice because there is no standard on how to build the ground-truth, each dataset was built using a different $\sigma$ value. The Gaussian filter can be written as:
\begin{equation}\label{eq:gau}
g(m,n)=\frac{1}{2\pi\sigma^2} \cdot e^{-\frac{m^2+n^2}{2\sigma^2}}
\end{equation}
where $m$ and $n$ represent the distance from the current location and $\sigma$ denotes the standard deviation. We search the $\sigma$ value used by the four saliency datasets, Toronto\cite{TORONTO}, MIT1003\cite{MIT1003}, CAT2000\cite{CAT2000} and SALICON \cite{SALICON}, based on the highest CC score achieved. As we can see in Table~\ref{tab:dataset}, the $\sigma$ value used varies across those datasets. A high performance can be achieved simply because the distributions of the training and test sets are similar and vice versa. We believe the location of ROI should be considered more for saliency instead of the distribution, a good model should capture the correct region but the shape (or contrast) is less important. One can imagine that all the distribution-based metrics suffer from the inherent drawback of shape sensitivity and it is difficult to avoid the center bias problem, see Section~\ref{sec:exp}.

\subsection{Location-based Metrics}
Location-based metrics do not rely on the distribution built by the Gaussian process(Equation~\ref{eq:gau}). Similar to CC, NSS is the recommended metric due to its equal penalty on FPs and FNs \cite{Bylinskii19}. However, NSS essentially considers all the fixation locations as positive and the others as negative, more FPs will be introduced when a larger $\sigma$ is applied on the training set. Our experiment validates this hypothesis by showing the highest NSS score is achieved when training on the setting of the smallest $\sigma$, see Section~\ref{sec:sigma}. This bias of NSS also makes it challenging to evaluate models across datasets.

The family of AUC metrics was criticized for ignoring FPs with small values \cite{Bylinskii19}. But the FPs will be ignored only when its value is smaller than the smallest threshold (value at fixations). That is, in practice, the ignored FPs would be \emph{relatively} small and this relative saliency is considered to be more important than absolute magnitudes \cite{Bruce15}. One study \cite{Bylinskii19} has shown that AUC metrics are robust to $\sigma$, but this happens only when the highest value of prediction is a TP. Our experiment shows the AUC metric is also slightly affected by the choice of $\sigma$, Section~\ref{sec:sigma}, but they are relatively more robust than CC and NSS. Nevertheless, the AUC metric is still the most promising way to overcome the center bias issue because we can directly sample negatives rather than considering distribution properties. Before discussing the center bias and demonstrating our method, we first show a potential problem in computing AUC metrics and our proposed global smoothing function in the next section.

\subsection{Global Smoothing Strategy}\label{sec:global}
As shown in Figure~\ref{fig:sample}, there are different ways to represent the fixation ground-truth. Both of the maps can be considered as matrices, the distribution map can also be interpreted as a 2D probability function and the fixation map can be converted into a set of positive locations. In this work, we demonstrate our method using all interpretations of saliency. For clarity, we denote a matrix as $\vec{X}$ or $\vec{Y}$ (an image or the fixation map, the first and second graphs in Figure~\ref{fig:sample}), a set of coordinates as $\mathcal{P}$ or $\mathcal{N}$ and the probability function as $f_{\vec{X}}$ (density or distribution maps, third graph in Figure~\ref{fig:sample}) and they can be converted to each other, see Section~\ref{sec:fnauc}.

All the AUC-based metrics are computed by applying various thresholds on the map to draw an ROC curve. This can be problematic when different positive locations share the same magnitude value. Let's denote an output map as $\hat{\vec{Y}}$ and we quantize the map into three value degrees to demonstrate this problem, $\{0, 0.5, 1\}$. It is worth noting that this problem is not artificial since CNNs tend to produce highly polarized and quantized output values that are often close to a binary map. One naive solution\footnote{According to the MIT benchmark \url{https://github.com/cvzoya/saliency/tree/master/code_forMetrics}} is to add a noise matrix $\vec{O}$ to the prediction map, $\hat{\vec{Y}} + \varepsilon \vec{O}$, $O_e \sim \mathcal{U}(0,1), \vec{O} \in R^{h \times w}$, where $O_i$ is $i$th element of the noise matrix and $\varepsilon$ is a small number. But this operation may break the relative saliency relationship due to the randomness.

\begin{figure}[!t]
	\setlength\tabcolsep{1.0pt}
	\begin{minipage}{0.5\textwidth}
	\centering
	\resizebox{0.8\textwidth}{!}{
	    \begin{tabular}{*{3}{c}}
		    \includegraphics[width=0.3\textwidth]{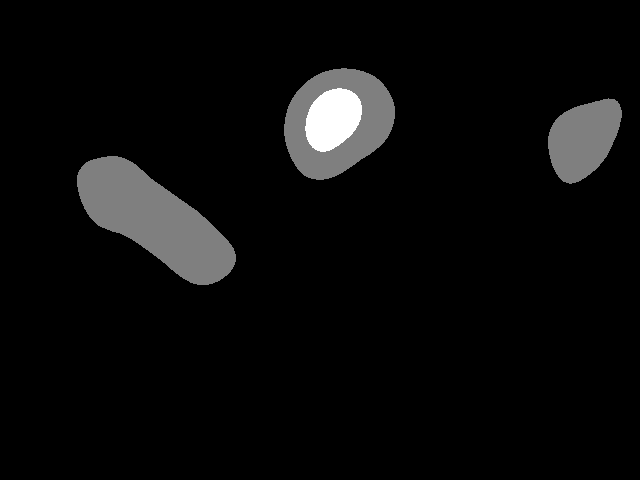}&
			\includegraphics[width=0.3\textwidth]{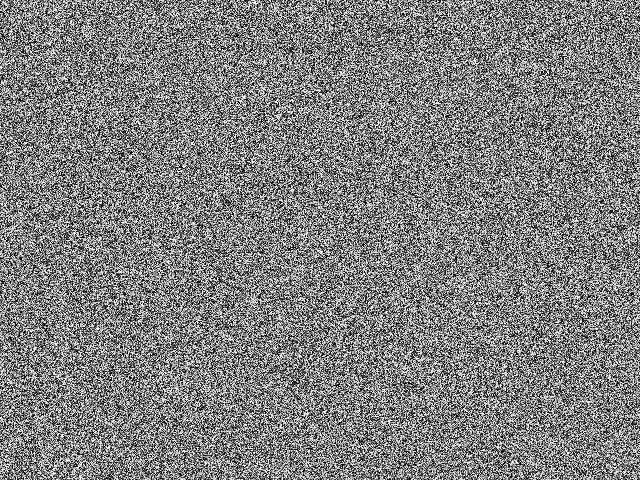}&
			\includegraphics[width=0.3\textwidth]{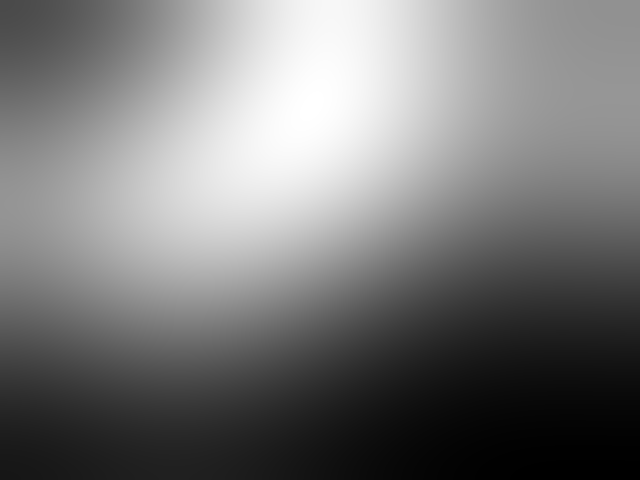}\\
			\small (a) Quantized & \small (b) Random Noise & \small (c) Global Gaussian 
		\end{tabular}}
		\caption{(a) A quantized example map to show the problem of limited value degrees. (b) A random noise map, $\vec{O}$, to jitter the output. (c) Our proposed global Gaussian map, $\vec{G}$.}
		\label{fig:global}
		\vspace{-0.1cm}
		\end{minipage}
	\begin{minipage}{0.5\textwidth}
	\centering
		\begin{tabular}{*{1}{c }}
		\includegraphics[width=0.8\textwidth]{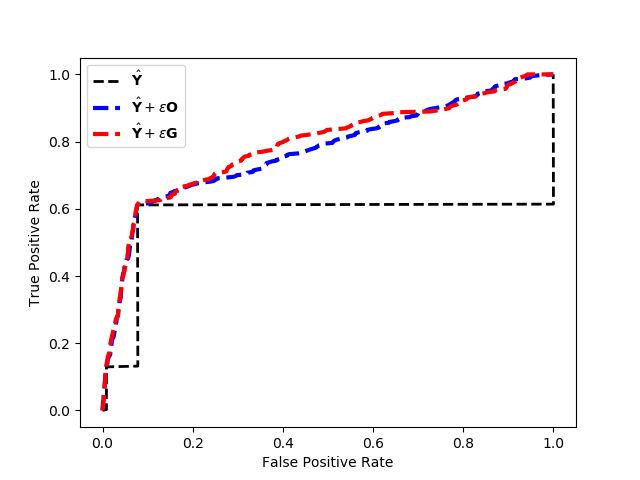}\\
		\end{tabular}
		\caption{ROC curves of the three maps, the AUC scores are: Quantized(black): 0.573, Jittered with Random Noise(blue): 0.774, Jittered with Global Gaussian(red): 0.790}
		\label{fig:gs_curve}
		\end{minipage}
		\vspace{-0.7cm}
	\end{figure}

In this work, we propose a global smoothing process to solve this problem. Instead of introducing random noise, we build a global Gaussian map, $\vec{G}$, by applying Equation~\ref{eq:gau} using a relatively large standard deviation to \emph{diversify} the value range, e.g., $\sigma=min(h/4, w/4)$. As shown in Figure~\ref{fig:global}, the noise matrix $\vec{O}$ used by \cite{MIT300} is similar to white Gaussian noise. Our global Gaussian map, $\vec{G}$, covers most areas of the image for tie-breaking. From Figure~\ref{fig:gs_curve}, we can see that our proposed map $G$ achieves a larger AUC comparing with the random noise $O$. Note that this improvement is model-agnostic in evaluation, but we are hoping it results in a more meaningful prediction for a fair model comparison. This operation can be also combined with the widely used Gaussian post-process, which utilizes a small value of $\sigma$ to achieve ``local'' smoothness. Given the small performance differences between models, this operation may arguable help to disambiguate which are in fact the best performing.

\section{Center Bias and Spatial Metrics}
\subsection{Center Bias}\label{sec:cb}
The measurement of saliency has suffered from the center bias problem for a long time. There exist various causes for center bias, e.g., viewing strategy, initial orbital positions or motor bias. The main reason behind this may also be the photographer bias, humans tend to place the most interesting object or region near the center of an image \cite{Tseng09,Parkhurst02,Parkhurst03,Tatler05}. Therefore this tendency makes it difficult to show how good a saliency model is, a ``faked'' high performance may result from centrally biased methods. An early saliency study \cite{Parkhurst02} has shown the stimuli in an image, e.g., color, intensity and orientation, are important in guiding attention. They also showed a discrepancy that the predictions are uniformly distributed within an image while the fixations are more likely to be near the center. This leads to a hypothesis that the low-level visual features have an indirect effect on attention, while the resulting `objectness' is more significant \cite{Wolfgang08}. Later an alternative explanation to their work was proposed by \cite{Borji13a}, the good performance achieved is because the objectness corresponds more with the center bias. 

\begin{figure}[t]
	\setlength\tabcolsep{1.0pt}
	\centering
	\resizebox{.45\textwidth}{!}{
		\begin{tabular}{*{4}{c}}
			\includegraphics[width=0.112\textwidth]{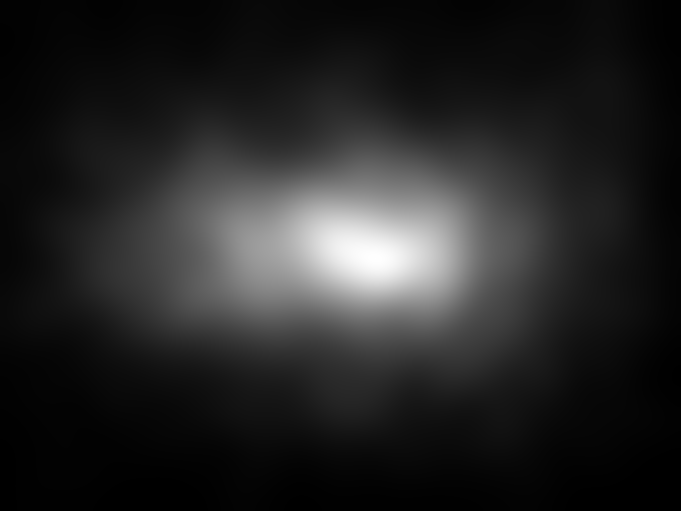}&
			\includegraphics[width=0.112\textwidth]{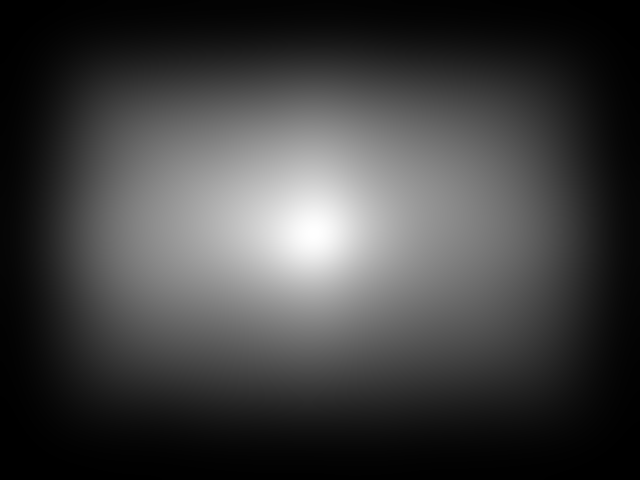}&
			\includegraphics[width=0.1495\textwidth]{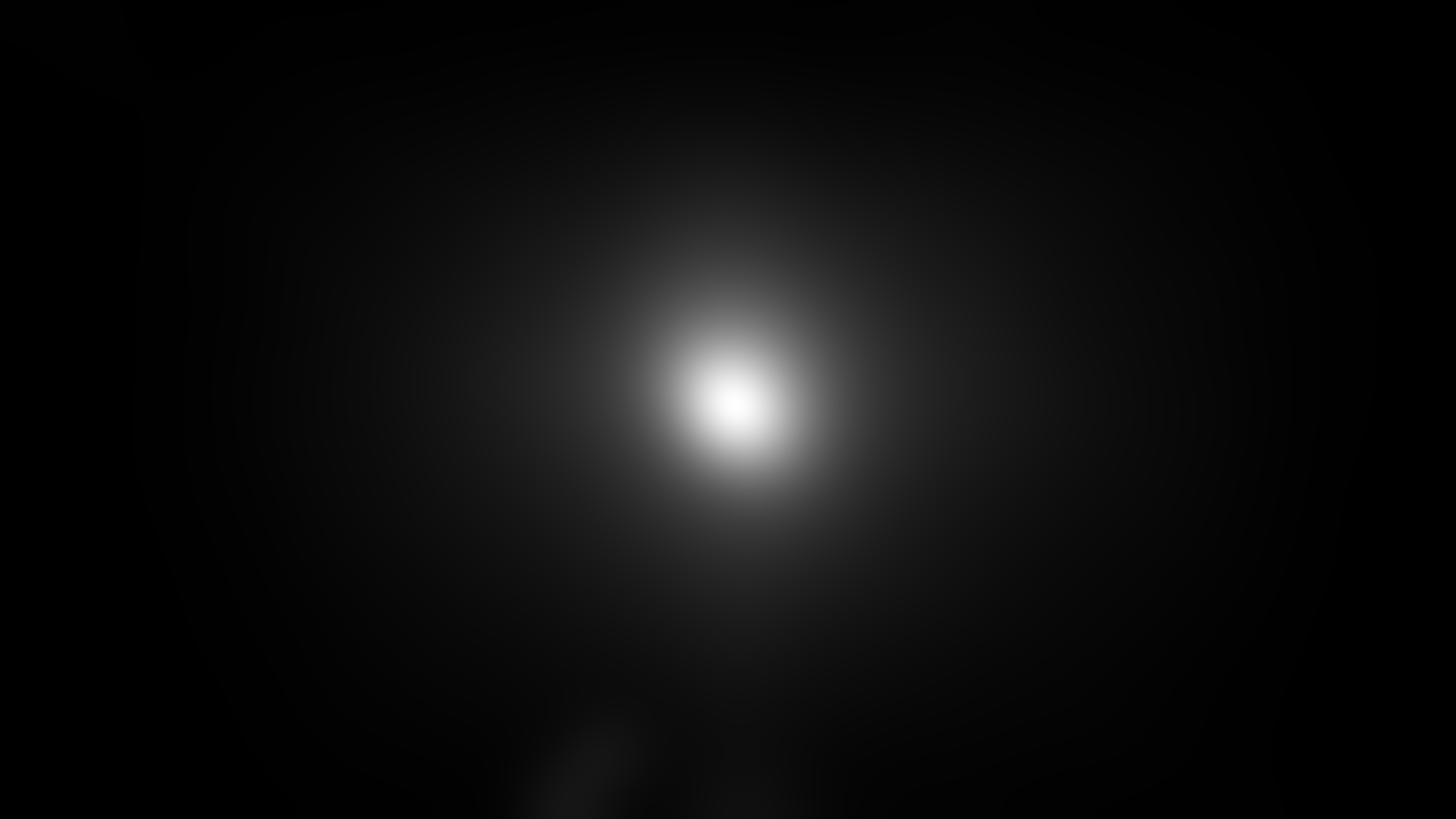}&
			\includegraphics[width=0.112\textwidth]{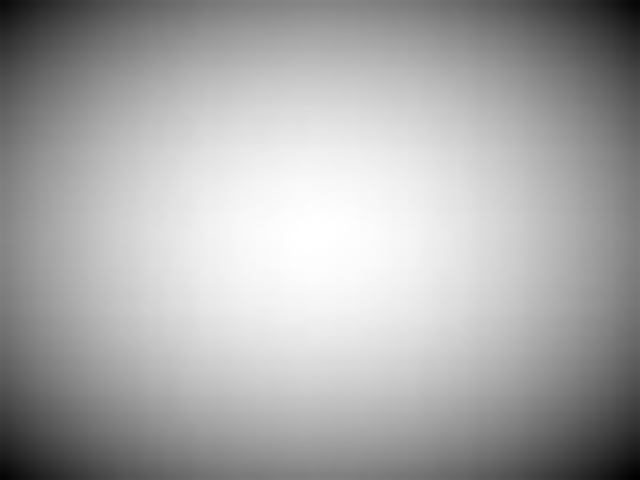}\\
			\small Toronto & \small SALICON & \small CAT2000 & \small Center Bias
		\end{tabular}}
		\caption{Distribution maps generated using all the fixations within each dataset and the center bias map from the MIT benchmark.}
		\label{fig:center}
		\vspace{-0.6cm}
\end{figure}

We show the distribution maps by applying Equation~\ref{eq:gau} on all the fixation locations within each dataset in Figure~\ref{fig:center}. The center bias is intrinsic to HVS across the datasets such that a synthetic center bias map, fourth map in Figure~\ref{fig:center} denoted as $\vec{CB}$, can achieve a decent performance by covering most of the fixation locations. We believe the objective of saliency prediction is to model the mechanism of HVS regardless of what types of features should be used or what the bias could be. The metric applied is expected to differentiate a good system from a synthetic map. Most of the metrics suffer from this problem because they are not designed for spatial biases, especially for the distribution-based case. In contrast, the location-based metric seems more promising on this issue.

The standard AUC metric was originally used for statistical analysis, and later was introduced to measure saliency performance by \cite{Bruce05}, also known as AUC-Judd \cite{MIT1003}, which considers all the non-fixated locations as negative. AUC-Borji \cite{AUCB} proposed to randomly sample negatives from all the non-fixated locations, which can be considered as a subset of AUC-Judd. But these two variants of AUC are not designed for the problem of center bias. S-AUC is a widely used metric specifically for the center bias, which samples negatives based on positive locations from other images within the same dataset. The assumption behind this is that the positives are also subject to a central Gaussian distribution so that they can be used to penalize the synthetic center map $\vec{CB}$, see Figure~\ref{fig:center}. But this sampling strategy ignores the spatial relationship between the positive and negative, which may result in an ``over-penalty'' to TPs. Furthermore, S-AUC may favor an ``anti-center'' prediction, saliency methods \cite{ImageSig, SDSR} that are biased to peripheral regions and can achieve a higher S-AUC score \cite{Bruce15}. 

\subsection{Farthest-Neighbor AUC}\label{sec:fnauc}
Before we show our proposed FN-AUC, we first formalize AUC metrics in different representations and discuss their own focuses in evaluation. Let $\vec{X} \in R^{h \times w \times 3}$ and $\vec{Y} \in R^{h \times w}$ denote the input image and its fixation ground-truth (second image in Figure~\ref{fig:sample}) respectively. The fixation map $\vec{Y}$ can be converted into a set of fixation locations, denoted as $\mathcal{F}=\{(m,n):\vec{Y}(m,n)=1,m=1 \dots w, n=1 \dots h \} $. The set of all the possible locations $\mathcal{S}$ can be formulated as $\mathcal{S}=\{(m,n):m=1 \dots w, n=1 \dots h\}$, $\mathcal{F} \subseteq \mathcal{S}$. When computing an AUC score, the positive set is $\mathcal{P} = \mathcal{F}$, but the generation of the negative set varies depending on the AUC metric applied. For AUC-Judd, it considers all the non-fixated locations as negative, $\mathcal{N}^{J}=\{\forall l \in \mathcal{S} : l \notin \mathcal{P}\} \iff \mathcal{S} \setminus \mathcal{P}$. For AUC-Borji, the negative set can be considered as applying $Bernoulli$ sampling on $\mathcal{N}^{J}$ with a cardinality constraint, $\mathcal{N}^{B} = \{s \in \mathcal{N}^{J} : |\mathcal{N}^{B}|=|\mathcal{P}|\}, \mathcal{N}^{B} \subseteq \mathcal{N}^{J}$ (a bijective function could be applied). It is obvious that both the AUC metrics actually focus on the same statistical property, but they can not overcome the center bias problem because no spatial information is utilized. 

For S-AUC, we build the positive set for the entire dataset by $\mathcal{P}_{all} = \{\mathcal{F}_1, \mathcal{F}_2, \dots, \mathcal{F}_N\}$, assuming there are $N$ samples in the dataset. The positive set of S-AUC is the same as AUC-Judd and AUC-Borji, $\mathcal{P}=\mathcal{F}, \mathcal{P} \subseteq \mathcal{P}_{all}$. For the negative set, S-AUC draws samples according to $\mathcal{N}^{S}=\{s \in \mathcal{P}_{all} : s \notin \mathcal{P}, |\mathcal{N}^{S}|=|\mathcal{P}|\}$. It is interesting to see that both S-AUC and AUC-Borji sample negatives from other sets, $\mathcal{P}_{all}$ and $\mathcal{N}^{J}$ respectively using the $Bernoulli$ sampling process. But S-AUC implicitly assumes that the set $\mathcal{P}_{all}$ is spatially subject to a centered Gaussian distribution (empirically validated by Figure~\ref{fig:center}) so that the sampled negatives can be used to penalize the center bias. But AUC-Judd and AUC-Borji do not make use of this spatial information.

Given the size of the image $(h, w)$, we can easily ``vectorize'' the total positive set $\mathcal{P}_{all}$ into a fixation map by first initializing a zero matrix, $\vec{Y_{all}}=\vec{0} \in R^{h \times w}$, then $\vec{Y_{all}}(m,n)=1 : \forall (m,n) \in \mathcal{P}_{all}$, let $v(\cdot)$ denote this ``vectorization'' conversion. A Gaussian filter (Equation~\ref{eq:gau}) is applied on the map $\vec{Y_{all}}$ to build the distribution map $f_{\vec{Y_{all}}}$ for each dataset as shown in Figure~\ref{fig:center}. Although the distribution map is a matrix, it can be interpreted as a 2D probability density function given the constraint  $\sum_{m=1}^{h}{\sum_{n=1}^{w}{{f_{\vec{Y_{all}}}}(m,n)}}=1$. In this way, the density maps in Figure~\ref{fig:center} can be viewed as the probability function of the S-AUC sampling process. Although each element in the negative set $\mathcal{N}^{S} \subseteq \mathcal{P}_{all}$ is drawn by $Bernoulli$ sampling, spatially it can be interpreted as a $Poisson$ sampling process such that the sampled elements have a higher probability to be located near the center (equal probability for sampling). 

Given the synthetic center bias map $\vec{CB}$ (as shown in Figure~\ref{fig:center}) and its probability function $f_{\vec{CB}}$, we can reformulate the center bias problem in terms of the distribution similarity between $f_{\vec{CB}}$ and $f_{\vec{Y_{all}}}$. The synthetic map is designed to ``mimic'' the distribution of the total fixations as a baseline such that the distance between the distributions should low or minimized, $arg\,min( d(f_{\vec{CB}}, f_{\vec{Y_{all}}}))$. The S-AUC metric can penalize the center bias because the probability distribution of $f_{v(\mathcal{N}^{S})}$ is similar to $f_{\vec{CB}}$, given that $\mathcal{N}^{S}$ is sampled from $\mathcal{P}_{all}$. 

\begin{figure}[t]
	\setlength\tabcolsep{1.0pt}
	\centering
	\resizebox{.45\textwidth}{!}{
		
			\includegraphics[width=1.\textwidth]{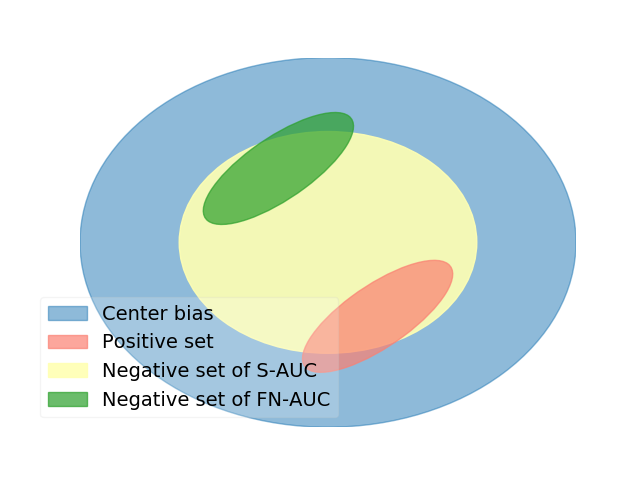}
			\vspace{-0.9cm}
	    }
		\caption{Diagram of our proposed FN-AUC vs S-AUC, our method aims at sampling a more directional negative set.}
		\label{fig:diagram}
		\vspace{-0.3cm}
\end{figure}

\begin{figure}[t]
	\setlength\tabcolsep{1.0pt}
	\centering
	\resizebox{.48\textwidth}{!}{
		\begin{tabular}{*{5}{c}}
			\includegraphics[width=0.2\textwidth]{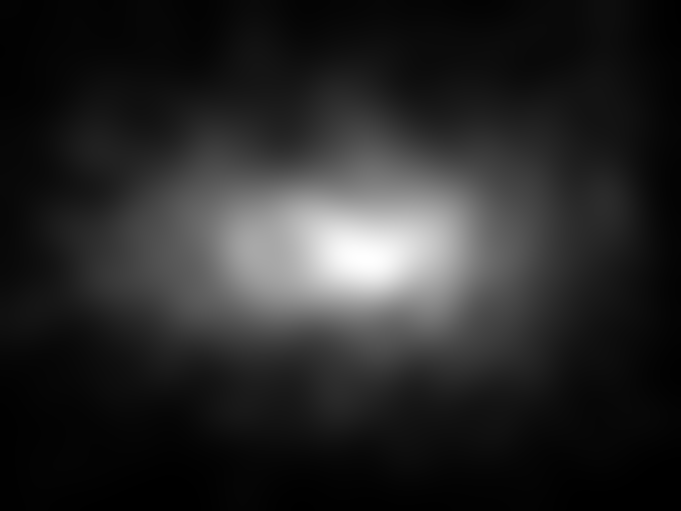}&
			\includegraphics[width=0.2\textwidth]{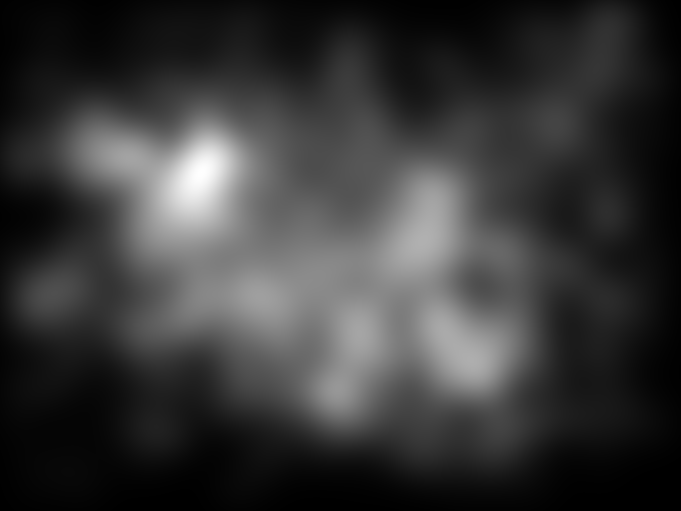}&
			\includegraphics[width=0.2\textwidth]{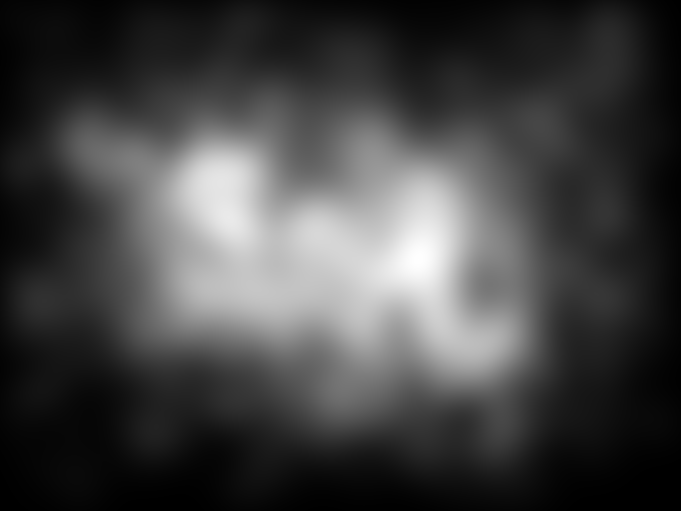}&
			\includegraphics[width=0.2\textwidth]{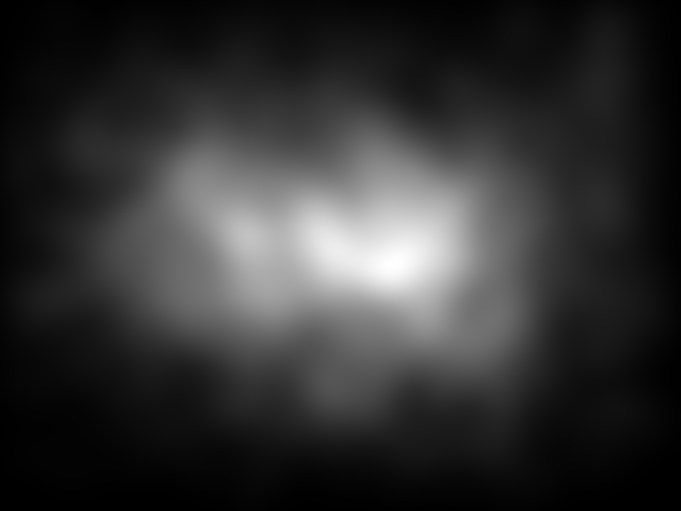}&
			\includegraphics[width=0.2\textwidth]{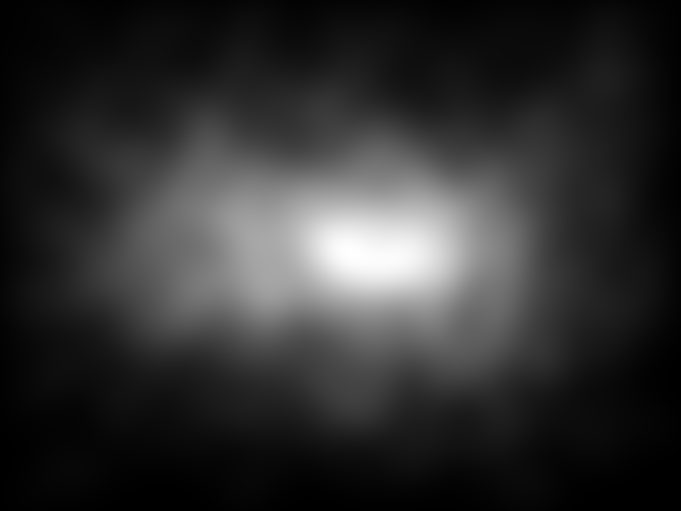}\\
            \Large S-AUC &\Large FN-AUC(K=5) &\Large FN-AUC(K=20) &\Large FN-AUC(K=50) &\Large FN-AUC(K=70) \\
        \end{tabular}}
		\caption{Distribution maps of the negative set sampled by S-AUC and FN-AUC, with different numbers of neighbors on the Toronto dataset.}
		\label{fig:negmap}
		\vspace{-0.6cm}
\end{figure}

However, the S-AUC metric only considers global position information, the negative sample should be near the center, but it ignores the relative spatial relationship between the positive and negative. The positive set is also a subset of the total $\mathcal{P} \subseteq \mathcal{P}_{all}$, which means spatially the probability function of $f_{v(\mathcal{N}^{S})}$ also overlaps with $f_{v(\mathcal{P})}$. This may lead to an over-penalty on the TP rate and also explains why S-AUC blindly favors peripherally-focused methods \cite{Bruce15}. To solve this problem, we propose to not only make use of the global information, but also take the relative spatial relationship into account. The sampled negative set should be able to penalize the center bias map $\vec{CB}$ meanwhile without affecting the positive set. It is easy to formulate this constraint in the representation of a probability function, $arg\,max( d(f_{v(\mathcal{P})}, f_{v(\mathcal{N}^{FN})}))$. That is, the sampled negative set by FN-AUC is designed to be far apart from the positive locations. We visually show the relationship between the negative sets drawn by S-AUC and our method in Figure~\ref{fig:diagram}. The negative set of S-AUC is near the center, which overlaps with the positive. While our method intends to avoid the positive locations but still sample within the area of the synthetic map $\vec{CB}$. 

\begin{table}[h]
\vspace{-0.6cm}
\begin{algorithm}[H]
\caption{Farthest-Neighbor AUC}\label{algo:fnauc}
\hspace*{\algorithmicindent} \textbf{Input:} {$(\vec{X_i},\vec{Y_i})$, $i$th Data Sample in the dataset. $\mathcal{P}_{all} = \{\mathcal{F}_1,\mathcal{F}_2, \dots, \mathcal{F}_N\}$, a set contains all the fixation locations within the dataset.} \\
\hspace*{\algorithmicindent} \textbf{Output:}{$\mathcal{N}^{FN}_i$ the negative set for $i$th sample.} 
\begin{algorithmic}[1]
\State Initialize an empty list, denoted as $L$.
\For{$j=1$ to $N$}
\If {$i \neq j$}
\State $d_j = d(f_{v(\mathcal{F}_i)}, f_{v(\mathcal{F}_j)})$
\State add $(d_j, \mathcal{F}_j)$ to $L$.
\EndIf
\EndFor
\State Sort $L$ in descending order based on $d_j$.  \Comment{suppose $d(\cdot, \cdot)$ is a similarity measure.}
\State Add the associated fixation set to the $\mathcal{N}^{FN}_i$ based on the top $K$ elements in $L$, $\mathcal{N}^{FN}_i=\{\mathcal{F}_k : (d_k, \mathcal{F}_k) \in L, k=1 \dots K\}.$ 
\State return $\mathcal{N}^{FN}$
\end{algorithmic}
\end{algorithm}
\vspace{-0.7cm}
\end{table}

\begin{figure*}[t]
	\setlength\tabcolsep{1.0pt}
	\def\arraystretch{0.5}
	\centering
	\resizebox{1.\textwidth}{!}{
	    \begin{tabular}{*{8}{c }}
	       
	       \includegraphics[width=0.1\textwidth]{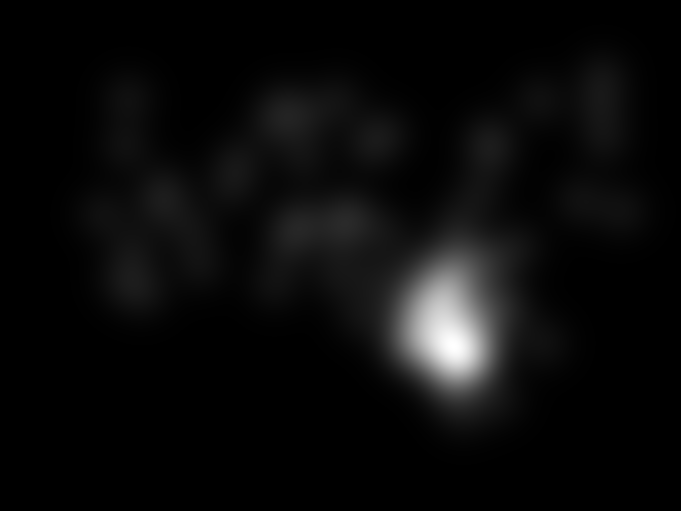}&
			\includegraphics[width=0.1\textwidth]{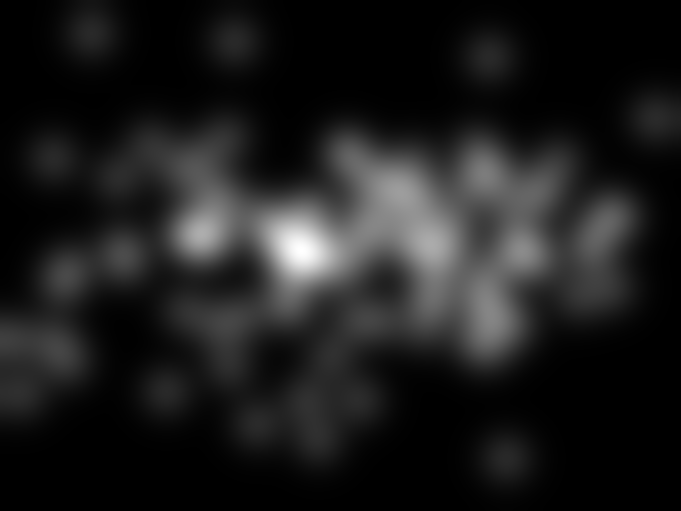}&
			\includegraphics[width=0.1\textwidth]{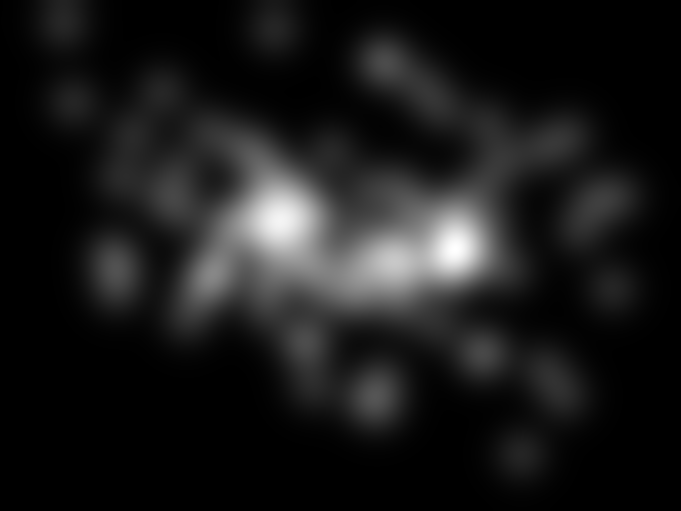}&
			\includegraphics[width=0.1\textwidth]{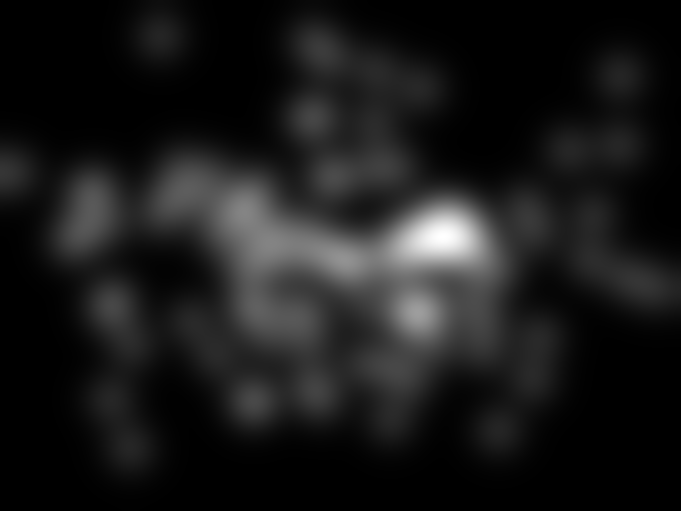}&
			\includegraphics[width=0.1\textwidth]{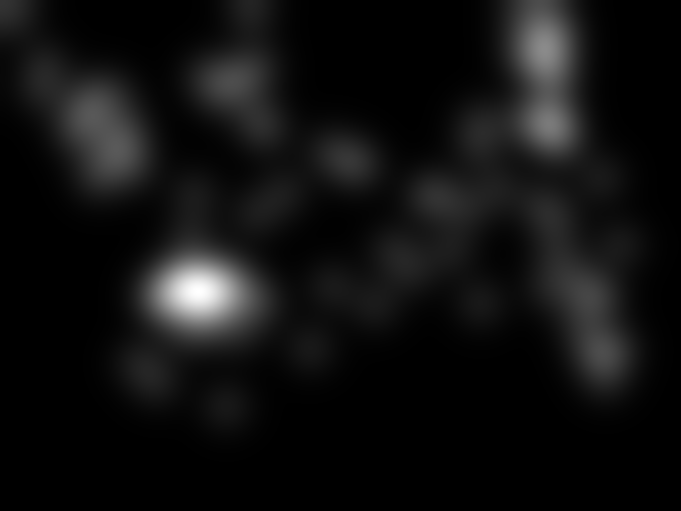}&
			\includegraphics[width=0.1\textwidth]{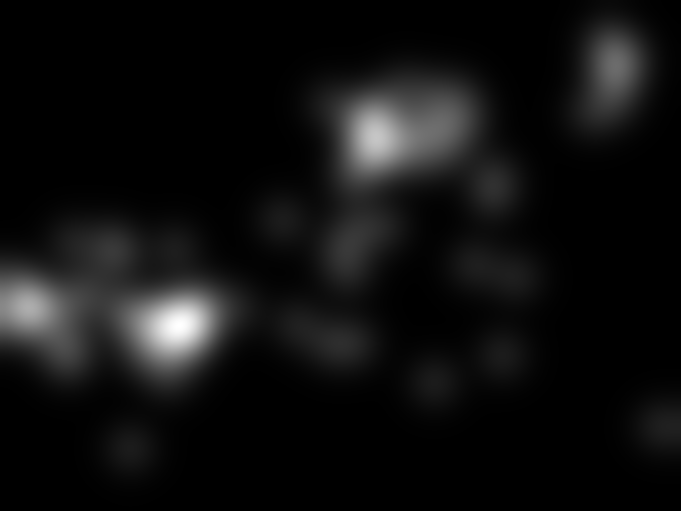}&
			\includegraphics[width=0.1\textwidth]{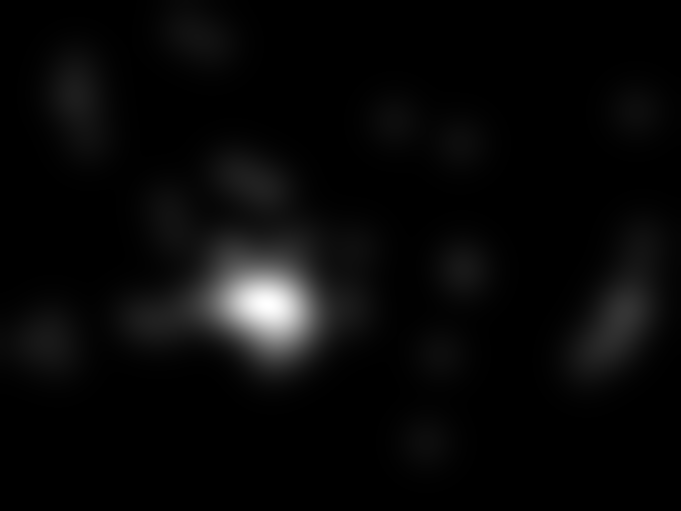}&
			\includegraphics[width=0.1\textwidth]{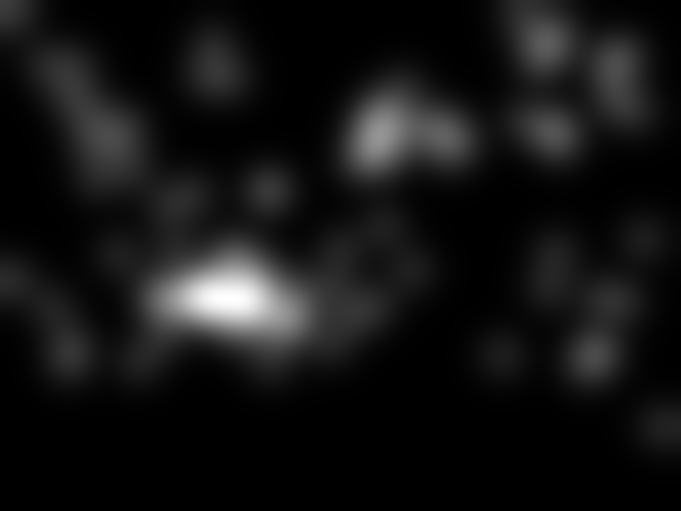}\\
			\tiny Pos & \tiny S(.769, .758, .987) & \tiny S(.812, .772, .951) & \tiny S(.779, .798, 1.025) & \tiny FN(.614, .552, .899) & \tiny FN(.648, .631, .974) & \tiny FN(.780, .689, .884) & \tiny Neg(.649, .494, .761)\\
			
			\includegraphics[width=0.1\textwidth]{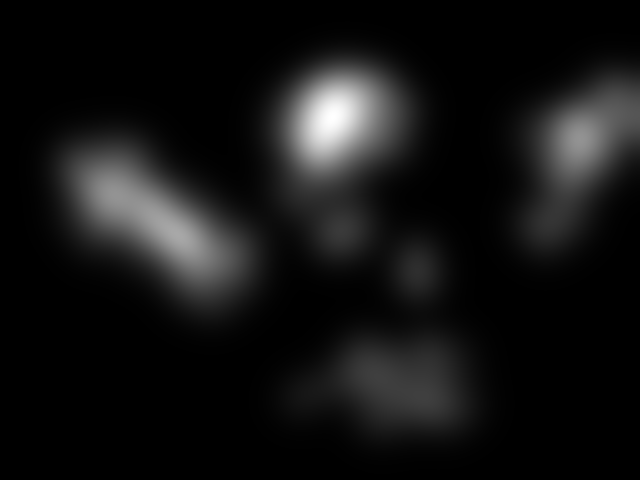}&
			\includegraphics[width=0.1\textwidth]{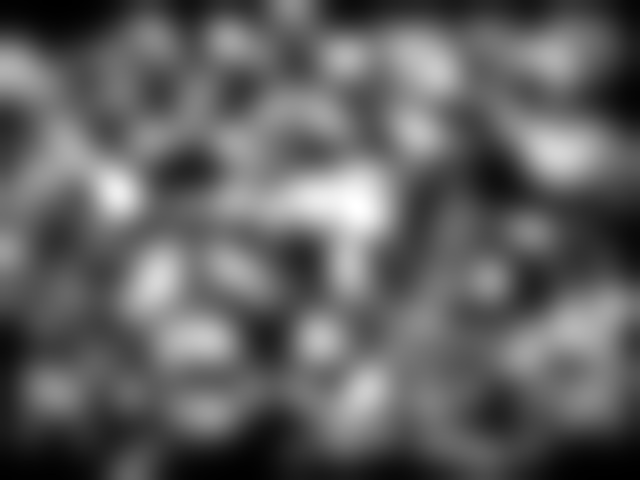}&
			\includegraphics[width=0.1\textwidth]{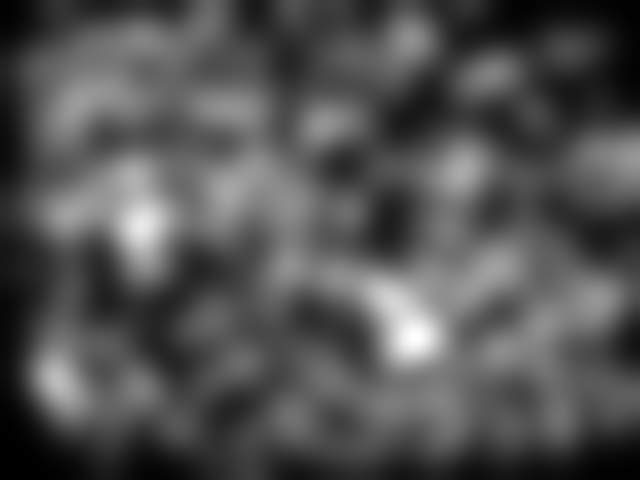}&
			\includegraphics[width=0.1\textwidth]{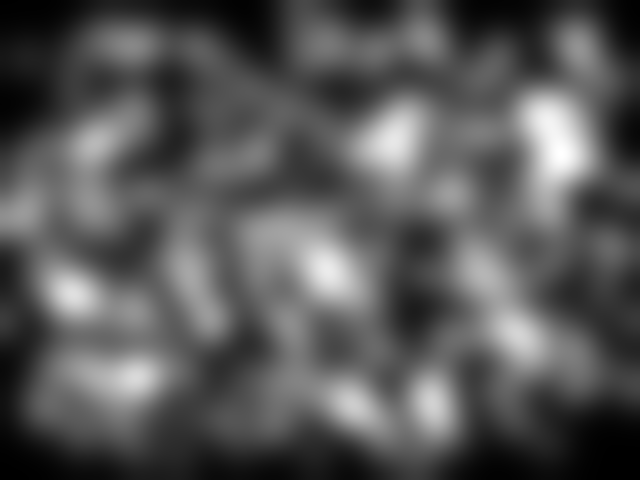}&
			\includegraphics[width=0.1\textwidth]{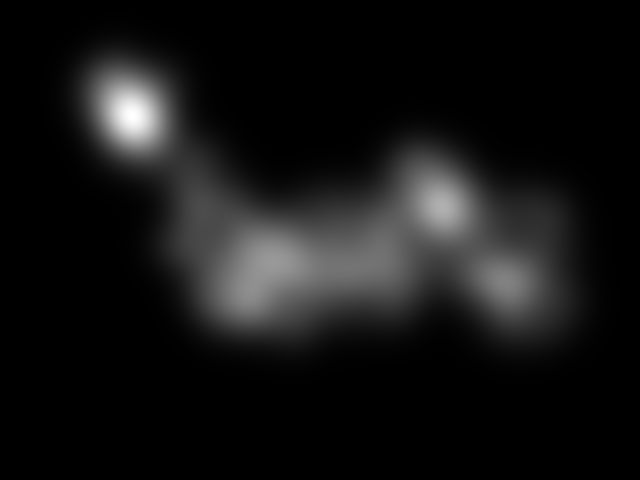}&
			\includegraphics[width=0.1\textwidth]{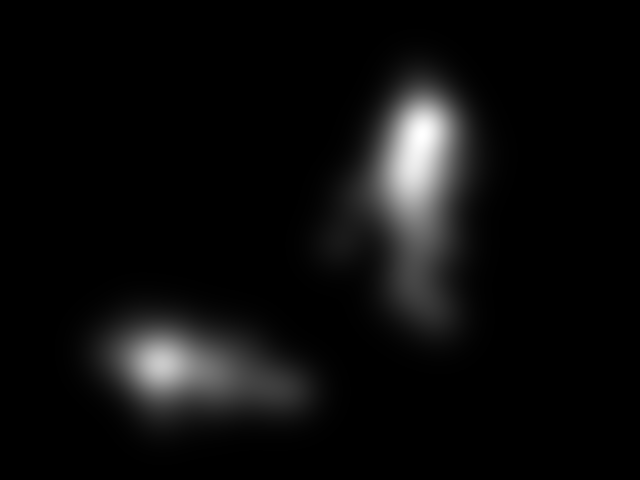}&
			\includegraphics[width=0.1\textwidth]{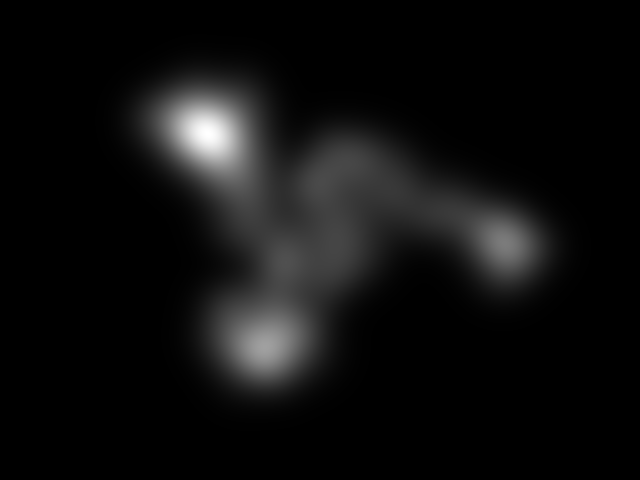}&
			\includegraphics[width=0.1\textwidth]{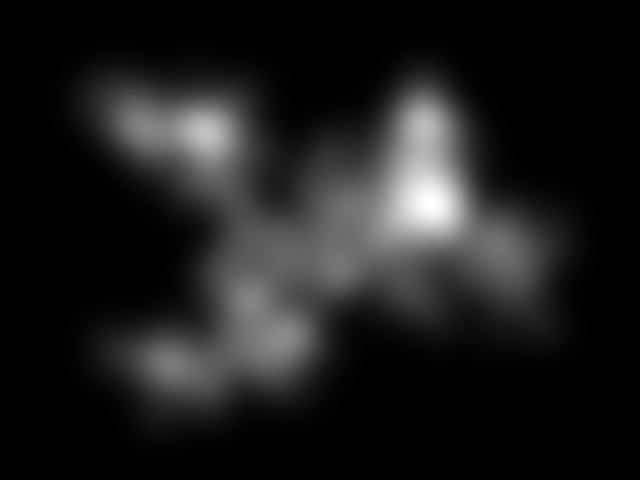}\\
			\tiny Pos & \tiny S(0.561, 0.687, 1.223) & \tiny S(0.567, 0.624, 1.102) & \tiny S(.578, .627, 1.085) & \tiny FN(0.791, 0.561, 0.710) & \tiny FN(0.751, 0.391, 0.520) & \tiny FN(0.829, 0.579, 0.698) & \tiny Neg(0.802, 0.529, 0.659)\\
	
		\end{tabular}}
		\caption{The negative set sampled by S-AUC and FN-AUC. The first column is the distribution map of positive locations. Columns $2\mbox{-}4$ are negative maps by S-AUC, column $5\mbox{-}7$ are negative maps by FN-AUC. The last column is the final output map of FN-AUC, sampled from columns $5\mbox{-}7$. The top row is a sample from the Toronto dataset drawn using Algorithm~\ref{algo:fnauc}, while the bottom is from the SALICON dataset drawn using the fast version of FN-AUC($K=3$, $CC<0$). The annotation represents ($\beta$, $\gamma$, $\gamma / \beta$).}
		\label{fig:fnsamples}
		\vspace{-0.5cm}
	\end{figure*}

We show how our FN-AUC samples the negative set $\mathcal{N}^{FN}$ in Algorithm~\ref{algo:fnauc}. The negative set consists of fixations from the neighbors that are least similar to the positive set, i.e. the farthest neighbors. Then we can sample from this negative set to have the same cardinality as the positive, $|\mathcal{N}^{FN}|=|\mathcal{P}|$. It is obvious that the FN-AUC sampling process has a complexity of $O(n)$. It is feasible to apply FN-AUC on a small dataset, e.g., Toronto, but it becomes problematic for large-scale datasets, e.g., SALICON. We also propose a fast version of FN-AUC for better scalability. Normally the number of fixations of each image is similar within one dataset. We can select only one farthest neighbor, $K=1$ in Algorithm~\ref{algo:fnauc} and omitting the cardinality constraint. More importantly, we can set an empirical threshold to select the first matched element without iterating over the entire dataset, e.g., a CC score below zero (inversely related). One extreme case could be that there exists one sample whose positive set is near the corner, every other sample may select it as the farthest neighbor such that FP rate is always zero. In this case, increasing the number of neighbors $K$ can deliver a more robust sampling process. However, we did not experience this problem on the datasets even applying $K=5$. When $K$ equals to the total number of the images within a dataset($K=N-1$), FN-AUC will reduce to S-AUC by sampling from $\mathcal{P}_{all}$. That is the larger value of $K$, the more we penalize the center bias map, we can see the effect of this trade-off of $K$ in Figure~\ref{fig:negmap}.

To compare with S-AUC, we propose to measure the quality of the negative set in two terms: 1) considering the negative set as positive locations and the center bias map $\vec{CB}$ as a prediction to measure the performance, hoping that a high score (e.g., CC or AUC) can be achieved so that the negative set can penalize $\vec{CB}$. 2) considering the negative set as a prediction map ($f_{v(\mathcal{N})}$) to measure its performance on the ground-truth $\mathcal{P}$, a low score is expected so that the negative set has little impact on the positive. Let's denote these two measures as $\uparrow \beta$ (the higher the better) and $\downarrow \gamma$ (the lower the better) respectively, we also show the ratio $\downarrow \gamma / \beta$ as an indicator of the quality. 

We can visually check the negative sets of $\mathcal{N}^S$ and $\mathcal{N}^{FN}$ in Figure~\ref{fig:fnsamples} (presented in distribution for visualization). Unsurprisingly, the random samples from S-AUC, columns $2\mbox{-}4$, tend to locate near the center, which leads to a higher $\beta$ value. But S-AUC dose not take the relative spatial relationship into account, the sampled negative map also largely overlaps with the positive, which results in a higher $\gamma$ value as well. While the three negative candidates drawn by FN-AUC(columns $5\mbox{-}7$) try to avoid penalizing the positive, therefore a lower $\gamma$ can be achieved. Moreover, the negative set still intersects with the center bias map, because the samples are drawn from $\mathcal{P}_{all}$ whose 2D distribution is similar to $f_{\vec{CB}}$. The final output negative set (the last column) sampled from the neighbors also achieves low scores of $\gamma$ and $\gamma / \beta$. It is interesting to see that our proposed FN-AUC has a more significant effect when evaluating on the SALICON dataset. The maps of S-AUC(bottom row) have ratio values of $\gamma / \beta > 1$, which indicates the negative set is penalized more on the positive over the center bias. The reason behind this is that the fixations of this dataset are more spread-out covering almost the whole image, see Figure~\ref{fig:center}. While our method can still generate more directional negative sets achieving low $\gamma / \beta$ values. (More examples are shown in Figure 1 in the supplementary material.)

\section{Experiment}\label{sec:exp}
\paragraph{Implementation Details:} SALICON is the largest saliency dataset. Its training set SALICON-train contains $10,000$ images with a resolution of $480 \times 640$ for training. We train CNN models on SALICON-train and report result on SALICON-val and the other datasets as shown in Table~\ref{tab:dataset}. The Toronto dataset has a similar image size to SALICON, so we simply resize the Toronto dataset during evaluation. But for MIT1003 and CAT2000, the ratio of the image size is completely different from SALICON. We apply the padding strategy used in \cite{SAMRes,MLNet}, each image has been resized and padded to keep the same ratio ($3/4$) as the input.

The network used is the ResNet-50 model, which has been pretrained on the ImageNet dataset. We simply apply the multi-level strategy used in \cite{MLNet,DEEPGAZE1} on the model, the side outputs from \{$conv1$, $conv10$, $conv22$, $conv40$, $conv49$\} are combined to generate the final prediction. The initial learning rate was set to $0.1$ with a weight decay of $1e^{-4}$. The total number of training epochs was $10$ and we reduced the learning rate every three epochs by multiplying by a factor of $0.1$. The batch size was set to $8$ and stochastic gradient descent was used to update the model after computing the mean squared error as the loss.

\vspace{-0.4cm}
\paragraph{Negative Set of FN-AUC:} To compute FN-AUC scores, we first build the negative set $\mathcal{N}^{FN}$ for each dataset. The metric of CC was used to compare the two distributions however other similarity measures can also be applied. To compare with S-AUC, we build a more directional negative set for FN-AUC in this experiment, $K=5$, and the final negative set is randomly drawn from the neighbors such that the number of elements is the same as the positive set. For Toronto, MIT1003 and CAT2000, we used the standard procedure as shown in Algorithm~\ref{algo:fnauc}. For SALICON, we applied the fast version of FN-AUC due to its large size with selected candidates chosen according to the first $K=5$ neighbors satisfying the requirement of $CC<0$. An optimal way to choose $K$ could be based on the average ratio of $\gamma / \beta$ across the entire dataset.

\subsection{The choice of $\sigma$} \label{sec:sigma}
As discussed in Section~\ref{sec:metric}, the distribution map of ground-truth varies according to the choice of $\sigma$. In this experiment, we show how this problem affects CNN-based systems when evaluating across datasets. Five different $\sigma$ values \{10, 20, 30, 40, 50\} are applied on SALICON-train to build ground-truth for training using Equation~\ref{eq:gau}. Then five CNN models with the same architecture are trained on each of the created ground-truth distributions. We evaluate the five models on different test sets using different metrics to show how sensitive those metrics are to the choice of $\sigma$. A metric can be considered \emph{biased} if a score for one model clearly outperforms or underperforms compared to the other models for the same metric since that the model architecture is the same. 

\begin{table}[t]
	    \centering
	        \resizebox{.48\textwidth}{!}{
				\begin{tabular}{cl|cccccc}
					\specialrule{1.2pt}{1pt}{1pt}\
					Dataset & Metric &$\sigma=10$& $\sigma=20$& $\sigma=30$ & $\sigma=40$ & $\sigma=50$ & Deviation \\
					\hline
					\multirow{6}{*}{\rotatebox[origin=c]{90}{\shortstack{\textbf{Toronto} \\ $\sigma=20$}}} 
					& CC & 0.684 & \textbf{0.694} & 0.684 & 0.669 & 0.647 & 0.016\\
					& NSS & \textbf{2.016} & 1.938 & 1.839 & 1.754 & 1.665 & 0.125\\
					& AUC-J & 0.852 & \textbf{0.856} & 0.855 & 0.853 & 0.854 & 0.001\\
					& AUC-B & 0.810 & 0.828 & 0.837 & 0.840 & \textbf{0.842} & 0.011\\
					& S-AUC & 0.713 & \textbf{0.717} & 0.713 & 0.705 & 0.694 & 0.008\\
					& FN-AUC & 0.805 & 0.817 & \textbf{0.824} & \textbf{0.824} & 0.820 & 0.006\\
					
					\hline
					\multirow{6}{*}{\rotatebox[origin=c]{90}{\shortstack{\textbf{CAT2000} \\ $\sigma=41$}}} 
					& CC & 0.539 & 0.556 & 0.556 & 0.558 & \textbf{0.559} & 0.007\\
					& NSS & 1.688 & \textbf{1.697} & 1.668 & 1.655 & 1.641 & 0.020\\
					& AUC-J & 0.846 & 0.853 & 0.854 & 0.859 & \textbf{0.864} & 0.005\\
					& AUC-B & 0.814 & 0.837 & 0.844 & 0.852 & \textbf{0.860} & 0.015\\
					& S-AUC & 0.702 & 0.716 & 0.719 & 0.725 & \textbf{0.732} & 0.010\\
					& FN-AUC & 0.704 & 0.716 & 0.716 & \textbf{0.719} & 0.717 & 0.005\\
					
					\hline
					\multirow{6}{*}{\rotatebox[origin=c]{90}{\shortstack{\textbf{MIT1003} \\ $\sigma=24$}}} 
					& CC & \textbf{0.623} & 0.610 & 0.586 & 0.558 & 0.532 & 0.033\\
					& NSS & \textbf{2.497} & 2.330 & 2.175 & 2.024 & 1.906 & 0.210\\
					& AUC-J & 0.897 & \textbf{0.899} & 0.898 & 0.897 & 0.895 & 0.001\\
					& AUC-B & 0.860 & 0.873 & 0.879 & \textbf{0.880} & \textbf{0.880} & 0.007\\
					& S-AUC & 0.807 & \textbf{0.816} & \textbf{0.816} & 0.812 & 0.809 & 0.003\\
					& FN-AUC & 0.758 & 0.778 & \textbf{0.788} & \textbf{0.788} & 0.783 & 0.011\\
					
					\hline
					\multirow{6}{*}{\rotatebox[origin=c]{90}{\shortstack{\textbf{SALICON} \\ $\sigma=19$}}} 
					& CC & 0.843 & \textbf{0.863} & 0.860 & 0.839 & 0.812 & 0.018\\
					& NSS & \textbf{1.895} & 1.832 & 1.757 & 1.668 & 1.580 & 0.112\\
					& AUC-J & 0.858 & \textbf{0.859} & 0.857 & 0.853 & 0.848 & 0.004\\
					& AUC-B & 0.811 & 0.834 & 0.843 & \textbf{0.845} & 0.842 & 0.012\\
					& S-AUC & 0.781 & 0.799 & \textbf{0.805} & 0.804 & 0.799 & 0.008\\
					& FN-AUC & 0.833 & 0.857 & \textbf{0.868} & 0.867 & 0.861 & 0.012\\
				    \specialrule{1.2pt}{1pt}{1pt}
				\end{tabular}
			}
			\caption{CNN models trained on ground-truth maps built using different $\sigma$ values. The $\sigma$ value used by each test set is shown. The highest score of each row is highlighted in bold. The standard deviation shows how robust each metric is to the change of $\sigma$. (AUC-J for AUC-Judd and AUC-B for AUC-Borji.)}
			\label{tab:sigma_result}
    \vspace{-0.3cm}
\end{table}

\begin{table}[t]
	    \centering
	        \resizebox{.4\textwidth}{!}{
				\begin{tabular}{cc|cc|cc}
					\specialrule{1.2pt}{1pt}{1pt}\
					CC & 0.018 & NSS & 0.116 & AUC-J & 0.002 \\ 
					\hline
					AUC-B & 0.011 & S-AUC & 0.007 & FN-AUC & 0.008 \\
				    \specialrule{1.2pt}{1pt}{1pt}
				\end{tabular}
			}
			\caption{The average deviation for each metric across datasets.}
			\label{tab:sigma_result2}
    \vspace{-0.7cm}
\end{table}

We report the results of the five models in Table~\ref{tab:sigma_result}. We can see that a high score of CC tends to be achieved by the matched distribution when similar $\sigma$ values are applied. The NSS metric favors a small $\sigma$ value and thus, less FPs are produced (a more sparse prediction). This experiment validates our discussion in Section~\ref{sec:metric}. NSS is sensitive to the $\sigma$ value applied on the training set while CC relies more on the $\sigma$ difference between the training and test set. A potential risk that may further limit the practical use of distribution-based metrics is that interpolation operations (e.g., bi-linear) can also affect distribution properties. As shown in Table~\ref{tab:sigma_result}, the AUC metrics are also sensitive to the choice of $\sigma$. For the AUC metrics, AUB-Borji, S-AUC and FN-AUC, there exists a sampling process which may deliver slight randomness into evaluation. We report the deviation of each setting in the last column of Table~\ref{tab:sigma_result} and its average score across datasets in Table~\ref{tab:sigma_result2}. We can see that the sensitivity of AUC metric is relatively smaller than CC and NSS. Even though AUC-Judd seems correlated with $\sigma$ from Table~\ref{tab:sigma_result}, a low deviation score denotes that it does not show which setting has a clear advantage. Larger deviations of CC and NSS may result from their value range or the computation process and due to this, they may deliver false intuitions with respect to the quality of the model.

\subsection{Spatial Biases}
The inner workings of NNs are still under exploration but it is established that the features learned by NNs contain high-level objectness, therefore CNN-based methods may share the same spatial bias as discussed in \cite{Borji13a}. Our proposed metric aims at solving the spatial bias problem, so we compare different types of ``early'' visual features(low-level) as well as the center bias map (Figure~\ref{fig:center}) using our metric. The center bias map, $\vec{CB}$, is taken from the MIT benchmark. The traditional saliency methods for comparison include: Itti \cite{ITTI}, AIM \cite{TORONTO}, GBVS \cite{GBVS}, SUN \cite{SUN}, SDSR \cite{SDSR}, CAS \cite{CAS}, AWS \cite{AWS}, SWD \cite{SWD} and ImageSig(RGB) \cite{ImageSig}. It has been shown that those traditional methods utilize various low-level features \cite{Bruce15, Borji13a, SUN}, which leads to relatively different predictions. Some of the hand-crafted methods are considerably time consuming (e.g., CAS needs more than $20$ seconds to process each image due to its multi-scale design). Therefore we only focus on the smallest dataset, Toronto, for simplicity. We take the model ($\sigma=20$) from the last experiment as a CNN baseline because it is the closest to the default settings of Toronto and SALICON. The metrics can be roughly categorized into bias-tolerant (CC, NSS, AUC-J, AUC-B) and bias-sensitive (S-AUC, FN-AUC).

\begin{table}[t]
	    \centering
	        \resizebox{.45\textwidth}{!}{
				\begin{tabular}{l|cccccc}
					\specialrule{1.2pt}{1pt}{1pt}\
					Method & CC & NSS & AUC-J & AUC-B & S-AUC & FN-AUC \\
					\hline
					CB &  0.397 & 0.969 & 0.802 & 0.786 & 0.515 & 0.607 \\
					Itti \cite{ITTI} &  0.270 & 0.820 & 0.693 & 0.677 & 0.638 & 0.701 \\
					AIM \cite{TORONTO} & 0.312 & 0.896 & 0.725 & 0.720 & 0.659 & 0.725 \\
					GBVS \cite{GBVS} & 0.569 & 1.519 & 0.829 & 0.819 & 0.636 & 0.747 \\
					SUN \cite{SUN} & 0.215 & 0.650 & 0.665 & 0.652 & 0.610 & 0.654 \\
					SDSR \cite{SDSR} & 0.403 & 1.096 & 0.763 & 0.756 & 0.697 & 0.786 \\
					CAS \cite{CAS} & 0.449 & 1.271 & 0.781 & 0.768 & 0.688 & 0.781\\
					AWS \cite{AWS} & 0.466 & 1.341 & 0.787 & 0.775 & 0.707 & 0.789 \\
					SWD \cite{SWD} & 0.575 & 1.523 & 0.836 & \textbf{0.828} & 0.632 & 0.741\\
					ImageSig \cite{ImageSig} & 0.396 & 1.085 & 0.762 & 0.749 & 0.679 & 0.753 \\ 
				    CNN & \textbf{0.694} & \textbf{1.938} & \textbf{0.856} & \textbf{0.828} & \textbf{0.717} & \textbf{0.817} \\
			        \specialrule{1.2pt}{1pt}{1pt}
				\end{tabular}
			}
			\caption{Comparison of different saliency methods on the Toronto dataset.}
			\label{tab:cb_result}
    \vspace{-0.6cm}
\end{table}

\begin{figure}[h]
	\centering
	 \includegraphics[width=0.45\textwidth]{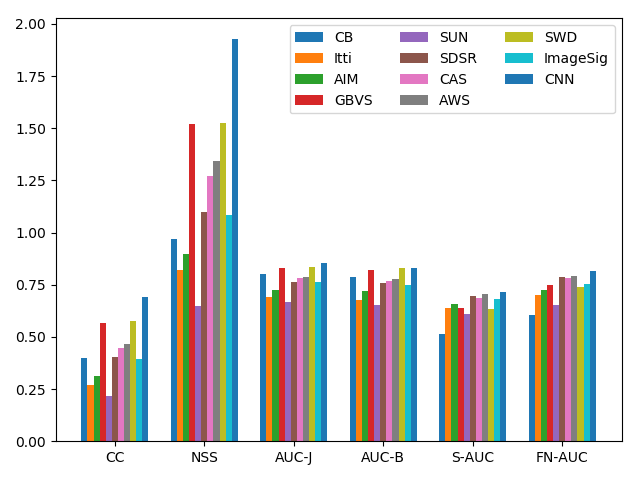}
	 \vspace{-0.2cm}
	\caption{Bar graph of the compared methods using different saliency metrics.}
		\label{fig:sota}
		\vspace{-0.3cm}
	\end{figure}

\begin{figure}[t]
	\setlength\tabcolsep{1.0pt}
	\centering
	\resizebox{.45\textwidth}{!}{
		\begin{tabular}{*{4}{c }}
            \includegraphics[width=0.17\textwidth]{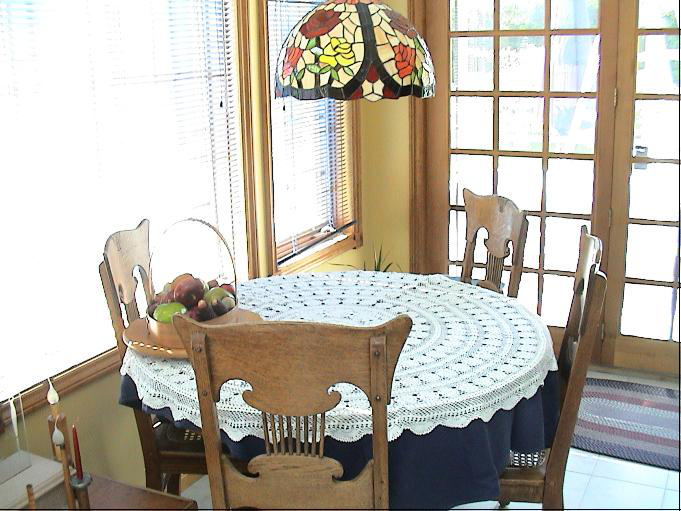}&
			\includegraphics[width=0.17\textwidth]{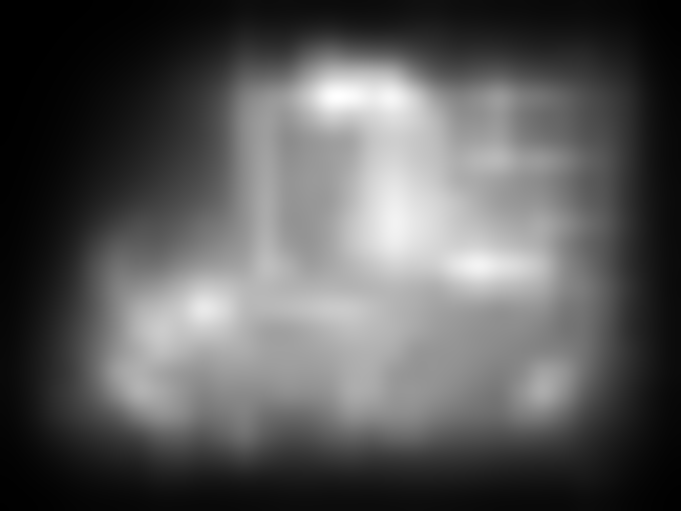}&
			\includegraphics[width=0.17\textwidth]{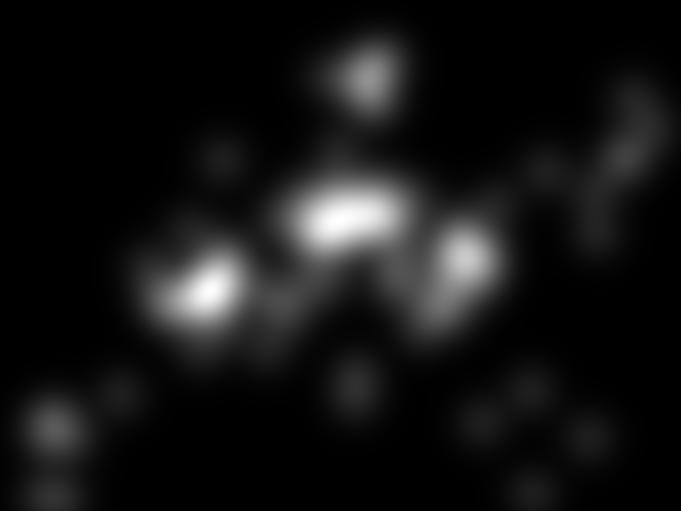}&
			\includegraphics[width=0.17\textwidth]{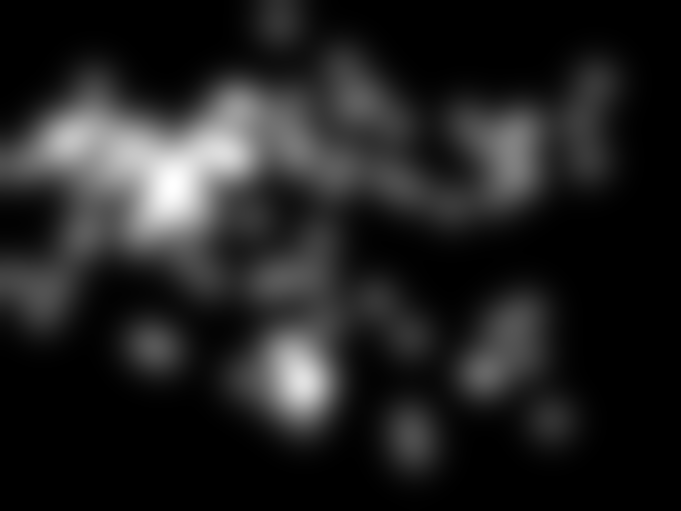}\\
			
		    \includegraphics[width=0.17\textwidth]{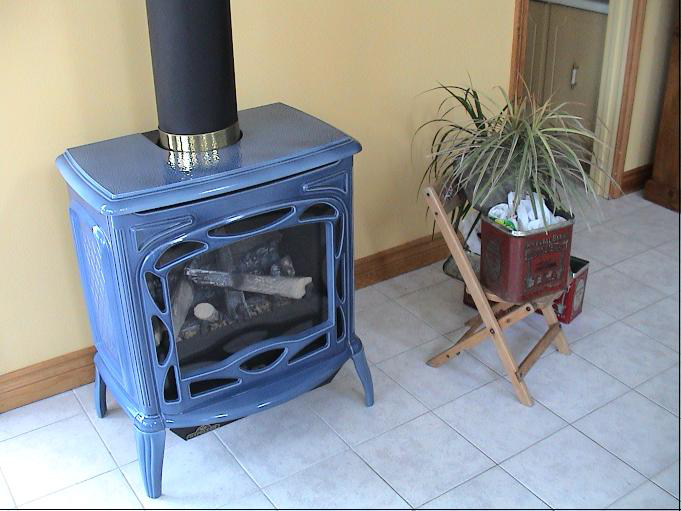}&
			\includegraphics[width=0.17\textwidth]{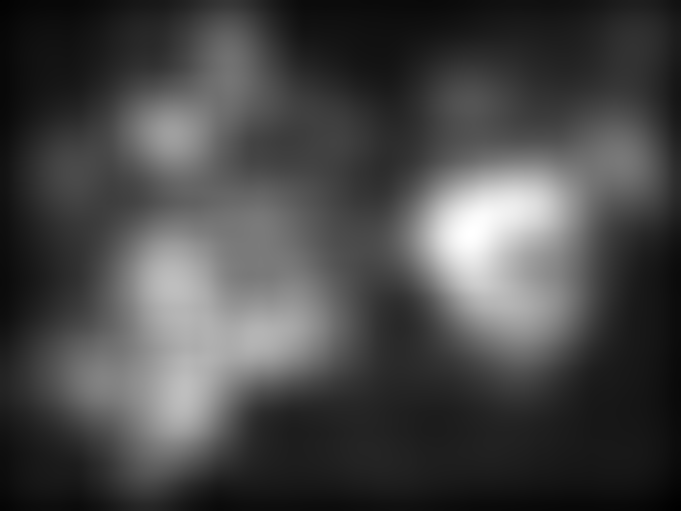}&
			\includegraphics[width=0.17\textwidth]{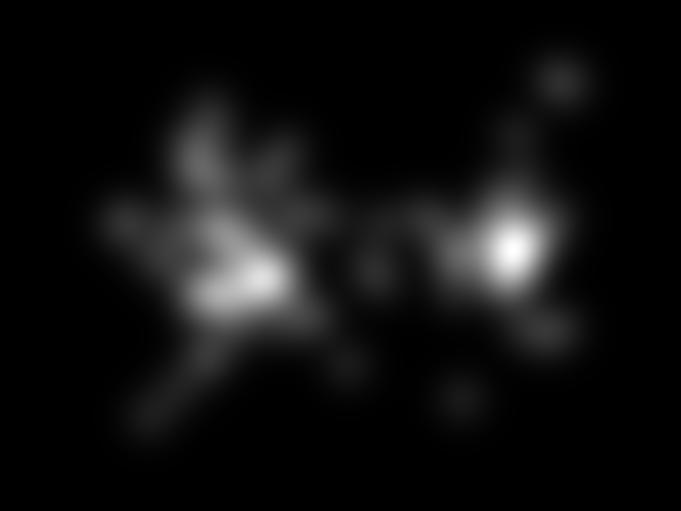}&
			\includegraphics[width=0.17\textwidth]{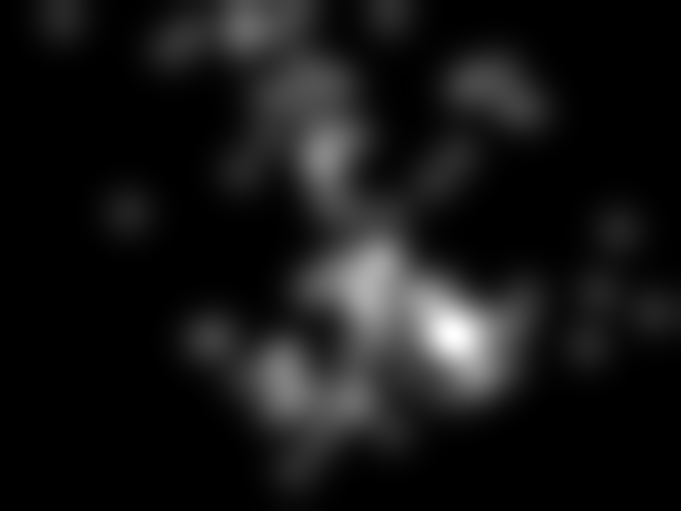}\\
			
			\includegraphics[width=0.17\textwidth]{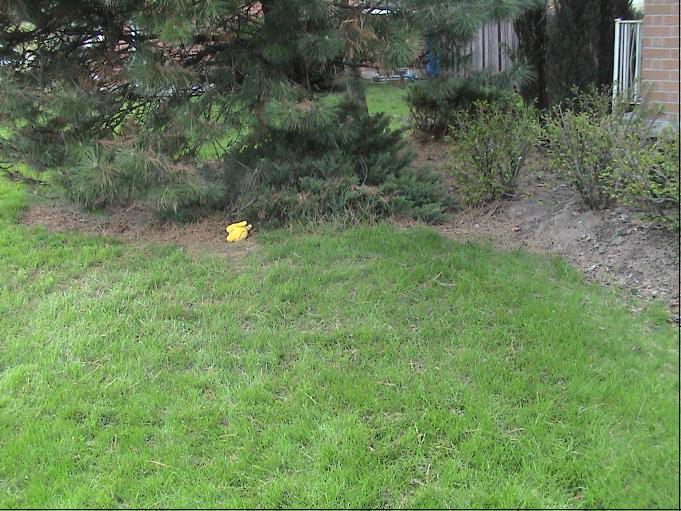}&
			\includegraphics[width=0.17\textwidth]{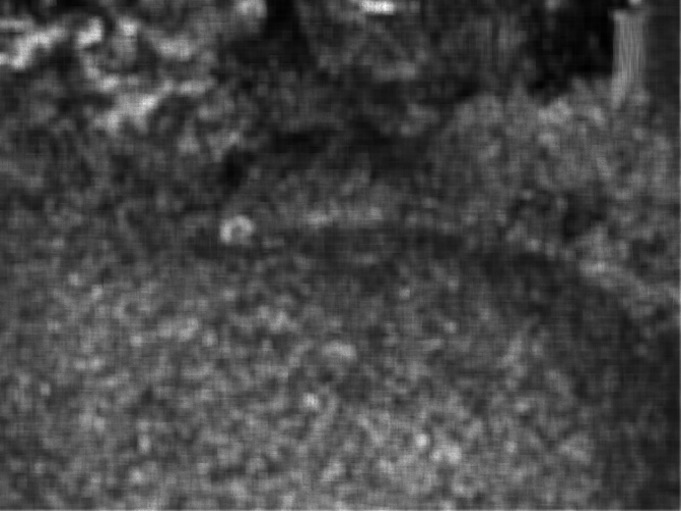}&
			\includegraphics[width=0.17\textwidth]{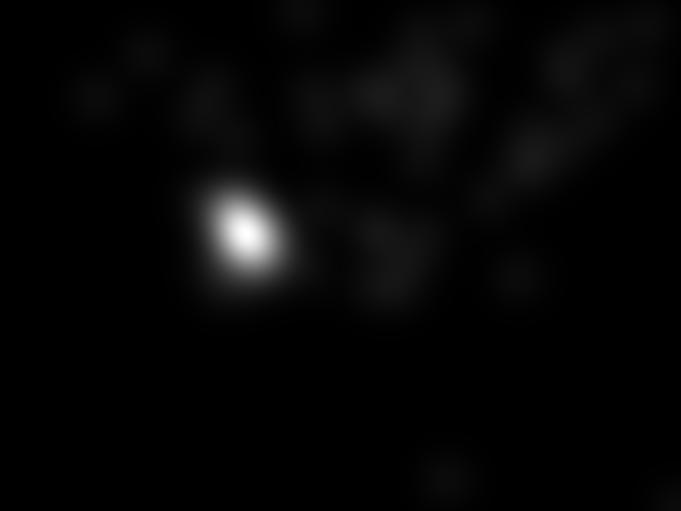}&
			\includegraphics[width=0.17\textwidth]{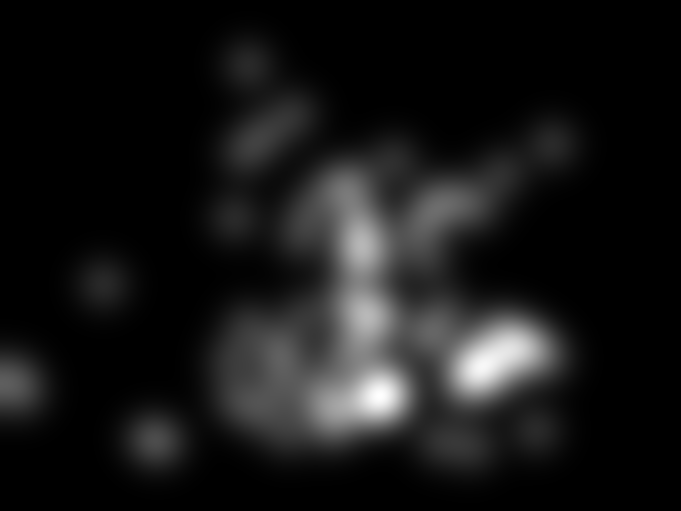}\\
			\includegraphics[width=0.17\textwidth]{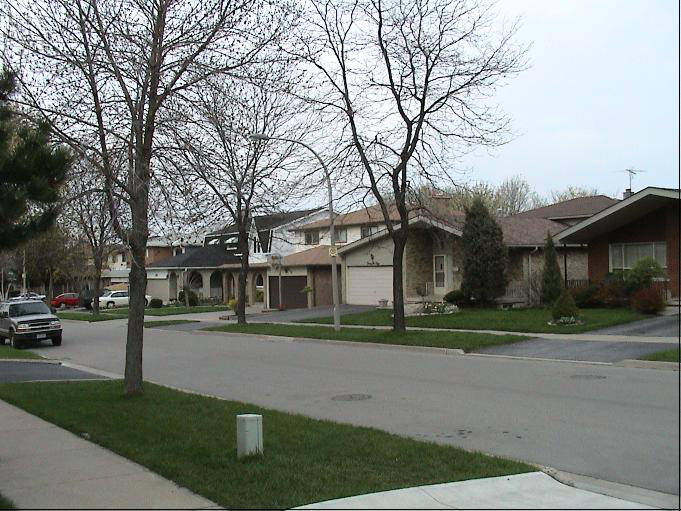}&
			\includegraphics[width=0.17\textwidth]{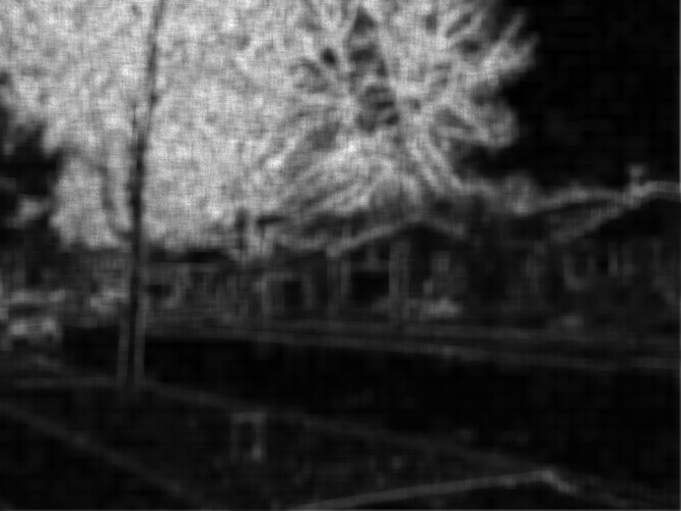}&
			\includegraphics[width=0.17\textwidth]{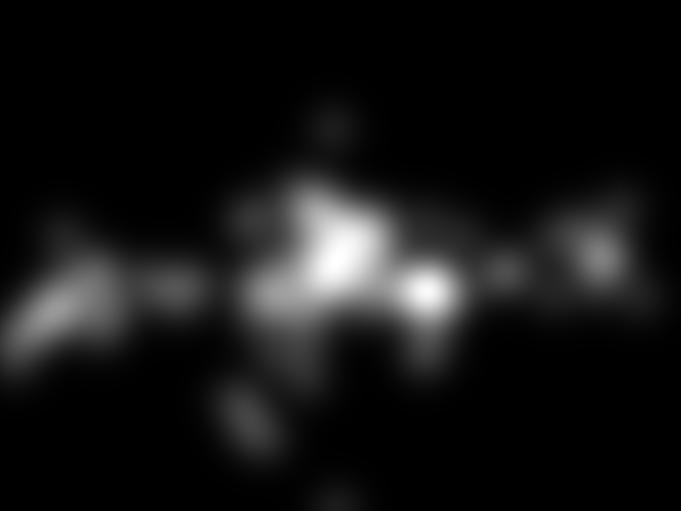}&
			\includegraphics[width=0.17\textwidth]{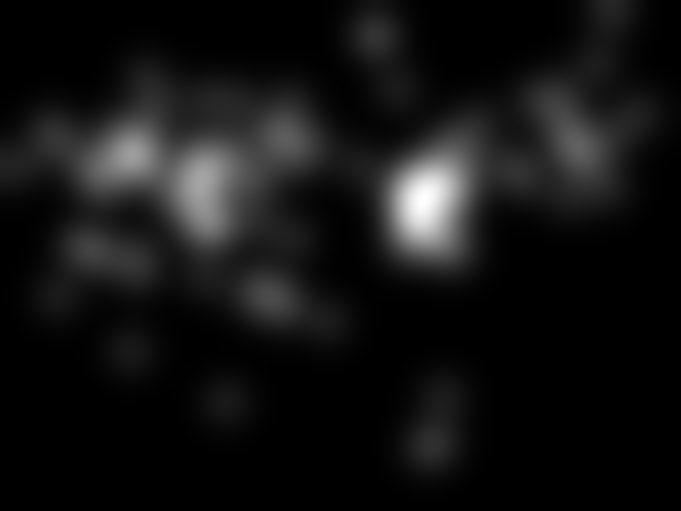}\\
			\large Image & \large Prediction & \large Fixation & \large FN-Negative
		\end{tabular}}
		\caption{Samples which have a large score difference between S-AUC and FN-AUC. From top to bottom (S-AUC, FN-AUC), 1:(0.573, 0.762), 2:(0.665, 0.867), 3:(0.540, 0.379), 4:(0.445, 0.285). The third column is the ground-truth distribution (positive) and the fourth column is the negative map drawn by FN-AUC.}
		\label{fig:case}
		\vspace{-0.7cm}
	\end{figure}

As shown in Table~\ref{tab:cb_result} and Figure~\ref{fig:sota}, we can see that the CNN model achieves the best performance on all the metrics, including FN-AUC. While it is trivial to compare high-level vs low-level features in this experiment, we are more interested in how the metrics measure the intrinsic spatial bias each method has. The first row, $\vec{CB}$, indicates a Gaussian map can achieve decent scores on most of the metrics. The $\vec{CB}$ map achieves higher FN-AUC ($0.607$) than S-AUC ($0.515$) because S-AUC only penalizes center bias as discussed in Section~\ref{sec:cb} (all the methods can outperform random guess $0.5$ on S-AUC). Moreover, Itti, AIM and SUN, all achieve lower performances on the bias-tolerant metrics, but our FN-AUC can still distinguish those methods from $\vec{CB}$. From the study \cite{SPROC}, the methods GBVS and AWS are the least and the most spatially consistent algorithms respectively. Therefore GBVS outperforms AWS on the bias-tolerant metrics, but AWS obtains higher scores on both of the bias-sensitive metrics, S-AUC and FN-AUC. We can also see this contrast between GBVS and other relatively consistent methods, SDSR, CAS and ImageSig. Moreover, our experiment shows the SWD method is even more spatially inconsistent than GBVS. When comparing with ``early'' vision, we are not surprised by the high performance achieved by the CNN due to its ability to learn high-level features. One thing that should be noted is although CNNs may tend to output center-favored maps due to its ``objectness'' knowledge, the CNN model still outperforms GBVS and SWD by a large margin. Both S-AUC and FN-AUC can penalize the spatial bias and we further investigate the difference between the two metrics in the next section.

\subsection{Case Discussion}\label{sec:case}
Our FN-AUC differs from S-AUC in that we also consider the relative relationship with the positive set. It is important to visually check the samples on which the metrics disagree with each other the most. We show samples which achieve a large score difference between S-AUC and FN-AUC in Figure~\ref{fig:case}. We can see from the scores, the top five rows achieve higher FN-AUC than S-AUC. From the third column, we can see that the ground-truth is near the center, but S-AUC will penalize the prediction regardless of whether it is a reasonable output or not. In contrast, our proposed FN-AUC achieves a higher score because the sampled negative is a more directional, rather than blind penalty. From the bottom four rows of Figure~\ref{fig:case}, we can see that the S-AUC score is higher than FN-AUC. We can see that the ground-truth is still near the center, but the predicted saliency region is near the periphery. In this case, those predictions should be considered as low-scoring outputs. But the negative set sampled by S-AUC will always be near the center as shown in Figure~\ref{fig:center} so that it cannot penalize the FPs. FN-AUC has a higher probability to penalize this type of prediction because it takes the spatial relationship into account. (More examples are shown in Figure 2 in the supplementary material.)

\section{Conclusion}
In this paper, we have shown that the NSS and CC metrics still suffer from sensitivity to the choice of the $\sigma$ value. This indicates that they can not fairly evaluate the CNN-based system on commonly used saliency datasets. NSS has been shown to be sensitive to the training set only, while CC is affected by the difference of $\sigma$ applied on the training and the test sets. The AUC metrics are relatively more robust to the change of $\sigma$. We delved into the AUC metrics based on different mathematical representations to show the drawback of S-AUC. Our proposed FN-AUC metric considers the relative position information so that a more directional negative set can be built to penalize the center bias only. Finally, our proposed global smoothing strategy can deliver a more stable AUC computation by retaining the saliency relationship. By no means can our method completely solve the problem of saliency evaluation, but our work sheds new light on the drawbacks of existing metrics as well as introduces a new sampling process. 

{\small
\bibliographystyle{ieee_fullname}
\bibliography{main}

\begin{thebibliography}{10}\itemsep=-1pt

\bibitem{Ancuti11}
C.~O. Ancuti, C. Ancuti, and P. Bekaert.
\newblock Enhancing by saliency-guided decolorization.
\newblock In {\em Proceedings of the 2011 IEEE Conference on Computer Vision
  and Pattern Recognition}, CVPR '11, pages 257--264, Washington, DC, USA,
  2011. IEEE Computer Society.

\bibitem{Boisvert16}
Jonathan Boisvert and Neil Bruce.
\newblock Predicting task from eye movements: On the importance of spatial
  distribution, dynamics, and image features.
\newblock {\em Neurocomputing}, 207, 05 2016.

\bibitem{CAT2000}
Ali Borji and Laurent Itti.
\newblock {CAT2000:} {A} large scale fixation dataset for boosting saliency
  research.
\newblock {\em CoRR}, abs/1505.03581, 2015.

\bibitem{Borji13a}
Ali Borji, Dicky~N. Sihite, and Laurent Itti.
\newblock {Objects do not predict fixations better than early saliency: A
  re-analysis of Einhäuser et al.'s data}.
\newblock {\em Journal of Vision}, 13(10):18--18, 08 2013.

\bibitem{AUCB}
A. {Borji}, D.~N. {Sihite}, and L. {Itti}.
\newblock Quantitative analysis of human-model agreement in visual saliency
  modeling: A comparative study.
\newblock {\em IEEE Transactions on Image Processing}, 22(1):55--69, Jan 2013.

\bibitem{Borji13b}
A. {Borji}, H.~R. {Tavakoli}, D.~N. {Sihite}, and L. {Itti}.
\newblock Analysis of scores, datasets, and models in visual saliency
  prediction.
\newblock In {\em 2013 IEEE International Conference on Computer Vision}, pages
  921--928, Dec 2013.

\bibitem{Bruce05}
Neil~D.B. Bruce and John~K. Tsotsos.
\newblock A statistical basis for visual field anisotropies.
\newblock {\em Neurocomputing}, 69(10):1301 -- 1304, 2006.
\newblock Computational Neuroscience: Trends in Research 2006.

\bibitem{Bruce15}
Neil~D.B. Bruce, Calden Wloka, Nick Frosst, Shafin Rahman, and John~K. Tsotsos.
\newblock On computational modeling of visual saliency: Examining what’s
  right, and what’s left.
\newblock {\em Vision Research}, 116:95 -- 112, 2015.
\newblock Computational Models of Visual Attention.

\bibitem{TORONTO}
Neil D.~B. Bruce and John~K. Tsotsos.
\newblock Saliency based on information maximization.
\newblock In {\em Proceedings of the 18th International Conference on Neural
  Information Processing Systems}, NIPS'05, pages 155--162, Cambridge, MA, USA,
  2005. MIT Press.

\bibitem{Bylinskii15}
Z. Bylinskii, E.M. DeGennaro, R. Rajalingham, H. Ruda, J. Zhang, and J.K.
  Tsotsos.
\newblock Towards the quantitative evaluation of visual attention models.
\newblock {\em Vision Research}, 116:258 -- 268, 2015.
\newblock Computational Models of Visual Attention.

\bibitem{Bylinskii19}
Z. {Bylinskii}, T. {Judd}, A. {Oliva}, A. {Torralba}, and F. {Durand}.
\newblock What do different evaluation metrics tell us about saliency models?
\newblock {\em IEEE Transactions on Pattern Analysis and Machine Intelligence},
  41(3):740--757, March 2019.

\bibitem{MLNet}
Marcella Cornia, Lorenzo Baraldi, Giuseppe Serra, and Rita Cucchiara.
\newblock A deep multi-level network for saliency prediction.
\newblock {\em CoRR}, abs/1609.01064, 2016.

\bibitem{SAMRes}
M. {Cornia}, L. {Baraldi}, G. {Serra}, and R. {Cucchiara}.
\newblock Predicting human eye fixations via an lstm-based saliency attentive
  model.
\newblock {\em IEEE Transactions on Image Processing}, 27(10):5142--5154, Oct
  2018.

\bibitem{SWD}
L. {Duan}, C. {Wu}, J. {Miao}, L. {Qing}, and Y. {Fu}.
\newblock Visual saliency detection by spatially weighted dissimilarity.
\newblock In {\em CVPR 2011}, pages 473--480, June 2011.

\bibitem{Wolfgang08}
Wolfgang Einhäuser, Merrielle Spain, and Pietro Perona.
\newblock {Objects predict fixations better than early saliency}.
\newblock {\em Journal of Vision}, 8(14):18--18, 11 2008.

\bibitem{Frintrop10}
Simone Frintrop, Erich Rome, and Henrik~I. Christensen.
\newblock Computational visual attention systems and their cognitive
  foundations: A survey.
\newblock {\em ACM Trans. Appl. Percept.}, 7(1):6:1--6:39, Jan. 2010.

\bibitem{AWS}
A. Garcia-Diaz, X.R. Fdez-Vidal, X.M. Pardo, and R. Dosil.
\newblock Saliency from hierarchical adaptation through decorrelation and
  variance normalization.
\newblock {\em Image and Vision Computing}, 30(1):51--64, 2012.

\bibitem{CAS}
S. {Goferman}, L. {Zelnik-Manor}, and A. {Tal}.
\newblock Context-aware saliency detection.
\newblock {\em IEEE Transactions on Pattern Analysis and Machine Intelligence},
  34(10):1915--1926, Oct 2012.

\bibitem{GBVS}
Jonathan Harel, Christof Koch, and Pietro Perona.
\newblock Graph-based visual saliency.
\newblock In {\em Proceedings of the 19th International Conference on Neural
  Information Processing Systems}, NIPS'06, pages 545--552, Cambridge, MA, USA,
  2006. MIT Press.

\bibitem{ImageSig}
X. {Hou}, J. {Harel}, and C. {Koch}.
\newblock Image signature: Highlighting sparse salient regions.
\newblock {\em IEEE Transactions on Pattern Analysis and Machine Intelligence},
  34(1):194--201, Jan 2012.

\bibitem{Hu18}
Yuan{-}Ting Hu, Jia{-}Bin Huang, and Alexander~G. Schwing.
\newblock Unsupervised video object segmentation using motion saliency-guided
  spatio-temporal propagation.
\newblock {\em CoRR}, abs/1809.01125, 2018.

\bibitem{ITTI}
L. {Itti}, C. {Koch}, and E. {Niebur}.
\newblock A model of saliency-based visual attention for rapid scene analysis.
\newblock {\em IEEE Transactions on Pattern Analysis and Machine Intelligence},
  20(11):1254--1259, Nov 1998.

\bibitem{PDP}
Saumya Jetley, Naila Murray, and Eleonora Vig.
\newblock End-to-end saliency mapping via probability distribution prediction.
\newblock {\em CoRR}, abs/1804.01793, 2018.

\bibitem{Jia17b}
S. Jia, Y. Zhang, D. Agrafiotis, and D. Bull.
\newblock Blind high dynamic range image quality assessment using deep
  learning.
\newblock In {\em 2017 IEEE International Conference on Image Processing
  (ICIP)}, pages 765--769, Sept 2017.

\bibitem{SALICON}
Ming Jiang, Shengsheng Huang, Juanyong Duan, and Qi Zhao.
\newblock Salicon: Saliency in context.
\newblock In {\em The IEEE Conference on Computer Vision and Pattern
  Recognition (CVPR)}, June 2015.

\bibitem{MIT300}
Tilke Judd, Frédo Durand, and Antonio Torralba.
\newblock A benchmark of computational models of saliency to predict human
  fixations.
\newblock 01 2012.

\bibitem{MIT1003}
T. {Judd}, K. {Ehinger}, F. {Durand}, and A. {Torralba}.
\newblock Learning to predict where humans look.
\newblock In {\em 2009 IEEE 12th International Conference on Computer Vision},
  pages 2106--2113, Sep. 2009.

\bibitem{KL}
S. Kullback and R.~A. Leibler.
\newblock On information and sufficiency.
\newblock {\em Ann. Math. Statist.}, 22(1):79--86, 03 1951.

\bibitem{DEEPGAZE1}
Matthias K{\"u}mmerer, Lucas Theis, and Matthias Bethge.
\newblock Deep gaze i: Boosting saliency prediction with feature maps trained
  on imagenet.
\newblock {\em CoRR}, abs/1411.1045, 2014.

\bibitem{Matthias15}
Matthias K{\"u}mmerer, Thomas S.~A. Wallis, and Matthias Bethge.
\newblock Information-theoretic model comparison unifies saliency metrics.
\newblock {\em Proceedings of the National Academy of Sciences},
  112(52):16054--16059, 2015.

\bibitem{Matthias18}
Matthias K{\"u}mmerer, Thomas S.~A. Wallis, and Matthias Bethge.
\newblock Saliency benchmarking made easy: Separating models, maps and metrics.
\newblock In Vittorio Ferrari, Martial Hebert, Cristian Sminchisescu, and Yair
  Weiss, editors, {\em Computer Vision -- ECCV 2018}, pages 798--814, Cham,
  2018. Springer International Publishing.

\bibitem{DEEPGAZE2}
Matthias Kummerer, Thomas S.~A. Wallis, Leon~A. Gatys, and Matthias Bethge.
\newblock Understanding low- and high-level contributions to fixation
  prediction.
\newblock In {\em The IEEE International Conference on Computer Vision (ICCV)},
  Oct 2017.

\bibitem{Li18}
Shengxi Li, Mai Xu, Yun Ren, and Zulin Wang.
\newblock Closed-form optimization on saliency-guided image compression for
  hevc-msp.
\newblock {\em Trans. Multi.}, 20(1):155--170, Jan. 2018.

\bibitem{Parkhurst02}
Derrick Parkhurst, Klinton Law, and Ernst Niebur.
\newblock Modeling the role of salience in the allocation of overt visual
  attention.
\newblock {\em Vision Research}, 42(1):107 -- 123, 2002.

\bibitem{Parkhurst03}
Derrick Parkhurst and Ernst Niebur.
\newblock Scene content selected by active vision.
\newblock {\em Spatial vision}, 16:125--54, 02 2003.

\bibitem{NSS}
Robert~J. Peters, Asha Iyer, Laurent Itti, and Christof Koch.
\newblock Components of bottom-up gaze allocation in natural images.
\newblock {\em Vision Research}, 45(18):2397 -- 2416, 2005.

\bibitem{Riche13}
N. {Riche}, M. {Duvinage}, M. {Mancas}, B. {Gosselin}, and T. {Dutoit}.
\newblock Saliency and human fixations: State-of-the-art and study of
  comparison metrics.
\newblock In {\em 2013 IEEE International Conference on Computer Vision}, pages
  1153--1160, Dec 2013.

\bibitem{EMD}
Y. {Rubner}, C. {Tomasi}, and L.~J. {Guibas}.
\newblock A metric for distributions with applications to image databases.
\newblock In {\em Sixth International Conference on Computer Vision (IEEE Cat.
  No.98CH36271)}, pages 59--66, Jan 1998.

\bibitem{SDSR}
Hae~Jong Seo and Peyman Milanfar.
\newblock {Static and space-time visual saliency detection by
  self-resemblance}.
\newblock {\em Journal of Vision}, 9(12):15--15, 11 2009.

\bibitem{Tatler05}
Benjamin~W. Tatler, Roland~J. Baddeley, and Iain~D. Gilchrist.
\newblock Visual correlates of fixation selection: effects of scale and time.
\newblock {\em Vision Research}, 45(5):643 -- 659, 2005.

\bibitem{Tseng09}
Po-He Tseng, Ran Carmi, Ian G.~M. Cameron, Douglas~P. Munoz, and Laurent Itti.
\newblock {Quantifying center bias of observers in free viewing of dynamic
  natural scenes}.
\newblock {\em Journal of Vision}, 9(7):4--4, 07 2009.

\bibitem{Wang17}
Fei Wang, Mengqing Jiang, Chen Qian, Shuo Yang, Cheng Li, Honggang Zhang,
  Xiaogang Wang, and Xiaoou Tang.
\newblock Residual attention network for image classification.
\newblock {\em CoRR}, abs/1704.06904, 2017.

\bibitem{SPROC}
C. {Wloka} and J. {Tstotsos}.
\newblock Spatially binned roc: A comprehensive saliency metric.
\newblock In {\em 2016 IEEE Conference on Computer Vision and Pattern
  Recognition (CVPR)}, pages 525--534, June 2016.

\bibitem{SUN}
Lingyun Zhang, Matthew~H. Tong, Tim~K. Marks, Honghao Shan, and Garrison~W.
  Cottrell.
\newblock {SUN: A Bayesian framework for saliency using natural statistics}.
\newblock {\em Journal of Vision}, 8(7):32--32, 12 2008.

\bibitem{Zhang16}
W. {Zhang}, A. {Borji}, Z. {Wang}, P. {Le Callet}, and H. {Liu}.
\newblock The application of visual saliency models in objective image quality
  assessment: A statistical evaluation.
\newblock {\em IEEE Transactions on Neural Networks and Learning Systems},
  27(6):1266--1278, June 2016.

\bibitem{Zund13}
F. {Zünd}, Y. {Pritch}, A. {Sorkine-Hornung}, S. {Mangold}, and T. {Gross}.
\newblock Content-aware compression using saliency-driven image retargeting.
\newblock In {\em 2013 IEEE International Conference on Image Processing},
  pages 1845--1849, Sep. 2013.

\end{thebibliography}
}

\end{document}